\pdfoutput=1

\documentclass[11pt]{article}

\usepackage[preprint]{acl}

\usepackage{times}
\usepackage{latexsym}
\usepackage{verbatim}
\usepackage{amsmath}
\usepackage{lscape}
\usepackage{longtable}

\usepackage[T1]{fontenc}

\usepackage[utf8]{inputenc}

\usepackage{microtype}

\usepackage{inconsolata}

\usepackage{graphicx}
\usepackage{tabularx}
\usepackage{setspace}
\usepackage{algpseudocode}

\newcommand{\ul}[1]{\underline{#1}}

\newcommand\savg{S$_{AVG}$}
\newcommand\scls{S$_{CLS}$}
\newcommand\trand{S$_{T_{rand}}$}

\newcommand{\vcent}[1]{\begin{minipage}{1.2cm} \vspace{-2cm} {\small #1} \end{minipage}}

\title{Testing the assumptions about the geometry of sentence embedding spaces:\\ the cosine measure need not apply}

\author{
  \textbf{Vivi Nastase\textsuperscript{1}} \and
  \textbf{Paola Merlo\textsuperscript{1,2}}
\\
  \textsuperscript{1}Idiap Research Institute, Martigny, Switzerland \\
  \textsuperscript{2}University of Geneva, Swizerland
  \\
\texttt{vivi.a.nastase@gmail.com, Paola.Merlo@unige.ch}
}
\begin{document}
\maketitle

\begin{abstract}
Transformer models learn to encode and decode an input text, and produce contextual token embeddings as a side-effect. The mapping from language into the embedding space maps words expressing similar concepts onto points that are close in the space. In practice, the reverse implication is also assumed: words corresponding to close points in this space are similar or related, those that are further are not. 

Does closeness in the embedding space extend to shared properties for sentence embeddings? We present an investigation of sentence embeddings and show that the geometry of their embedding space is not predictive of their relative performances on a variety of tasks.

We compute sentence embeddings in three ways: as averaged token embeddings, as the embedding of the special [CLS] token, and as the embedding of a random token from the sentence. We explore whether there is a correlation between the distance between sentence embedding variations and their performance on linguistic tasks, and whether despite their distances, they do encode the same information in the same manner.

The results show that the cosine similarity -- which treats dimensions shallowly -- captures (shallow) commonalities or differences between sentence embeddings, which are not predictive of their performance on specific tasks. Linguistic information is rather encoded in weighted combinations of different dimensions, which are not reflected in the geometry of the sentence embedding space.


\end{abstract}

\section{Introduction}



Projecting words and larger pieces of text into an $n$-dimensional space allows us to map linguistic objects into a well-defined mathematical space, with specific metrics and operations. Building this projection relies on equating word similarity in language with closeness between their corresponding vectors in the embedding space, that is,  the embedding space is {\it smooth} \cite{bengio-ea2013}. The smoothness of the embedding space comes with several assumptions: similar or related words or sentences will be projected to points that are close in the space, and words or sentences corresponding to points that are close in the space are similar or related. Distance or similarity metrics in this space are the basis for the functioning of all current LLMs and their applications, and understanding the topology of the embedding space can bring insights both into the successes and also the failures of these models. 

Analysis of the embedding spaces  of the tokens through similarity measures have revealed that the spaces of many LLMs are anisotropic, with most words appearing in a narrow cone in this space, thus making distance metrics less informative \cite{timkey-van-schijndel-2021-bark,cai-etal-2021-isotropy}. 
These analyses are shallow: each dimension of these vectors is treated independently of the other dimensions. 
The dimensions may encode some information at a shallow level -- e.g. length, or extreme word frequencies within the sentence \cite{nikolaev-pado-2023-universe}; sentence structure through the words' relative positions \cite{Manning2020EmergentLS} -- but they do not correspond to linguistic features such as phrase type, or semantic role. The interplay among embedding dimensions is complex, as each can contribute to various linguistic features in different measures \cite{bengio-ea2013,elhage2022toymodelssuperposition}. This implies that the level at which words and sentence embeddings share features is not at a level that the cosine metric can detect. This explanation may also shed light on the apparent contradiction between the word embedding space being anisotropic \cite{mimno-thompson-2017-strange,timkey-van-schijndel-2021-bark,cai-etal-2021-isotropy} while word embeddings still leading to good results on a variety of NLP tasks (e.g. \cite{mercatali-freitas-2021-disentangling-generative,bao-etal-2019-generating,chen-etal-2019-multi}). 

We present an investigation of sentence embeddings and show that the geometry of their embedding space is not predictive of their relative performances on a variety of tasks. We consider four pretrained models from the BERT family
. For each model, we study and compare three variations of sentence representations: the averaged token embeddings, the embedding of the special [CLS] token, and a random token embedding.

While the vocabulary is finite, the sentences in a language are not. We therefore study the embedding space of sentences by comparing variations of representations for the same sentence, which should (theoretically) be very close as they should contain the same information. For this analysis, we start with the commonly used method --the cosine similarity-- and establish how close these representation variations are on a dataset of sentences extracted from the ParaCrawl corpus \cite{banon-etal-2020-paracrawl} (Section \ref{sec:sent_repr}). In the next step, we test these sentence representations on the FlashHolmes benchmark \cite{waldis-etal-2024-holmes}, which contains morphological, syntactic, semantic, discourse and reasoning tasks, and test whether embeddings that are close in the embedding space lead to similar performance on linguistic tasks  (Section \ref{sec:holmes}). Finally, we mine for sentence structure information, and test whether this kind of information is encoded consistently across the three representation variations of a sentence (Section \ref{sec:structure}).


The results show that closeness in the embedding space is not predictive of closeness of performance. In particular, for RoBERTa, all representation variations of a sentence are close in the embedding space, but their performance on the FlashHolmes tasks are very different. For Electra and DeBERTa, the \scls~ representations are almost orthogonal to the \savg~ representations, while having very close performance on the FlashHolmes tasks, and on deeper probing for syntactic structure. 

These seemingly contradictory results between distance in the embedding space and performance on various tasks support the view that the geometry of the embedding space is not a good proxy for investigating shared information among sentence embeddings. We discover linguistic features encoded through deeper combinations of weighted dimensions. 
Our work sheds lights on the embedding space, how linguistic information is encoded in embeddings and that cosine similarity is a shallow metric that does not provide information about (deep) shared features among words or sentences.


\section{Sentence representation comparisons in the embedding space}


We investigate the embedding space of sentences, and whether the common assumption -- that close points correspond to textual units that are similar -- is true for sentence representations. 

\subsection{Sentence representations}

\paragraph{Averaged token embeddings: \savg} The representation obtained by averaging a sentence's tokens (without the special [CLS] and [SEP] tokens) is frequently used as the  sentence's embedding \citep{nikolaev-pado-2023-investigating}. This representation benefits from the fact that the learning signal for transformer models is stronger at the token level, with a much weaker objective at the sentence level -- next sentence prediction \citep{devlin-etal-2019-bert,liu2019roberta}, sentence order prediction \citep{lan-etal-2020-albert}.

\paragraph{The embedding of the special [CLS] token: \scls} This representation is most commonly used after fine-tuning for specific tasks such as story continuation \citep{ippolito-etal-2020-toward}, sentence similarity \citep{reimers-gurevych-2019-sentence}, alignment to semantic features \cite{opitz-frank-2022-sbert}. 

\paragraph{The embedding of a random token: \trand} Using this as the sentence's representation can reveal how much contextual information each token embedding contains. 

\vspace{3mm}
\noindent We investigate these three variations of representing sentences in four pretrained transformer models: BERT\footnote{\url{https://huggingface.co/google-bert/bert-base-multilingual-cased}}, RoBERTa\footnote{\url{https://huggingface.co/FacebookAI/xlm-roberta-base}}, DeBERTa\footnote{\url{https://huggingface.co/microsoft/deberta-v3-base}}, and Electra\footnote{\url{google/electra-base-discriminator}}. BERT is the baseline transformer model. RoBERTa is a variation of BERT with optimized training, BPE tokenization, dynamic masking and without a next sentence objective \cite{liu2019roberta}. DeBERTA is another variation that introduces disentangled attention and an optimized mask decoder training \cite{he-etal-2021-deberta}. Unlike BERT, RoBERTa and DeBERTa, Electra is not a masked language model, rather implements a replaced token recognition model, predicting whether a token in the input was produced by a generator model \cite{clark2020electra}. Electra also outperforms XLNet \cite{yang2019xlnet} and MPNet \cite{song2020mpnet}.
The differences in the training regime and architecture of the models are reflected in the relative position of the embeddings in the embedding space (Section \ref{sec:sent_repr}).

\subsection{Analysis of the embedding space}

Our investigation starts from shallow analyses based on the cosine metric between variations of sentence embeddings, tests these embeddings on a benchmark covering different NLP tasks,
and probes for shared syntactic information, to establish whether closeness in the embedding space means close performance on NLP tasks.

\noindent
We first use the cosine metric to quantify how close the tokens in a sentence are to each other (Section \ref{sec:tokens}). According to the properties of the embedding space, this provides information about how similar the tokens are. Considering that most tokens in a sentence are not actually similar, and that the token embeddings are contextual, the cosine similarity rather quantifies how much contextual information these embeddings share.

\noindent
We then use the cosine metric to quantify the distance between the three sentence representation variations (Section~\ref{sec:sent_repr}). Since they represent the same sentence, \savg~ and \scls~ encode the same information, and the embedding space assumption dictates that they be close in the embedding space. \trand, as the encoding of only one token, encodes less information about the sentence, and is expected to be further apart in the embedding space from both \savg~ and \scls. We test whether these expectations are met.

\noindent
In the third step, we use each variation of a sentence representation to solve the FlashHolmes linguistic tasks. The goal is to verify whether relative distances in the embedding space -- quantified in the previous step -- are reflected in the relative performance of the three sentence embedding variations on the benchmark (Section~\ref{sec:holmes}).

\noindent
Finally, we test whether the three sentence representation variations all encode information about the chunk structure of a sentence, and if they do, whether it is encoded in the same way (Section~\ref{sec:structure}).

\subsection{Data}

The dataset consists of 1000 sentences in six languages (English, French, German, Italian, Romanian, Spanish) extracted from the parallel ParaCrawl corpus \cite{banon-etal-2020-paracrawl} (the datasets are not parallel)\footnote{The data will be made available upon publication.}. Each sentence is represented through the three representation variations.


\subsection{Contextual information in token and word embeddings}
\label{sec:tokens}

\begin{figure}[h]
    \begin{tabular}{cc}
    \begin{minipage}{0.47\linewidth}     
    {\small BERT}
    
    \includegraphics[width=\linewidth,trim={0 0 0 1cm},clip]{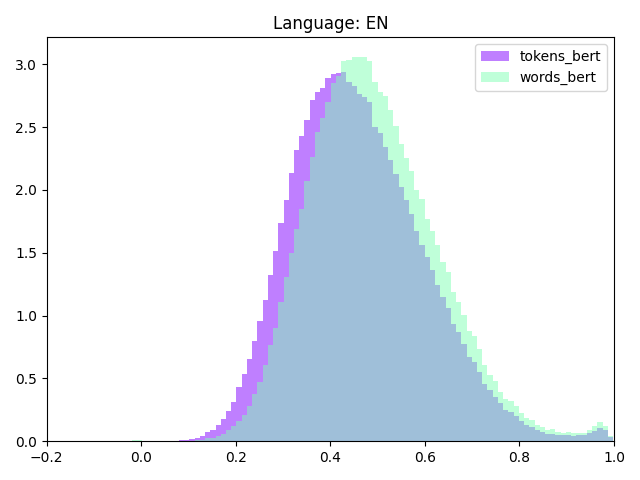}
    \end{minipage}
    &
    \begin{minipage}{0.47\linewidth}     
    {\small RoBERTa}
    
    \includegraphics[width=\linewidth,trim={0 0 0 0.9cm},clip]{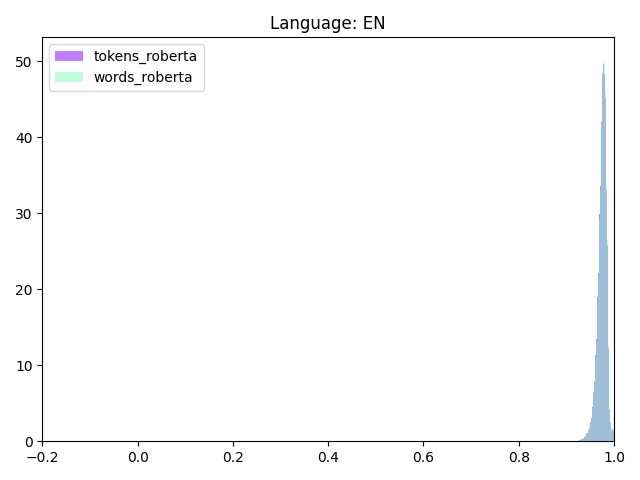}
    \end{minipage}
    
    \\ 
    
    \begin{minipage}{0.47\linewidth}
    {\small DeBERTa}
    
    \includegraphics[width=\linewidth,trim={0 0 0 1cm},clip]{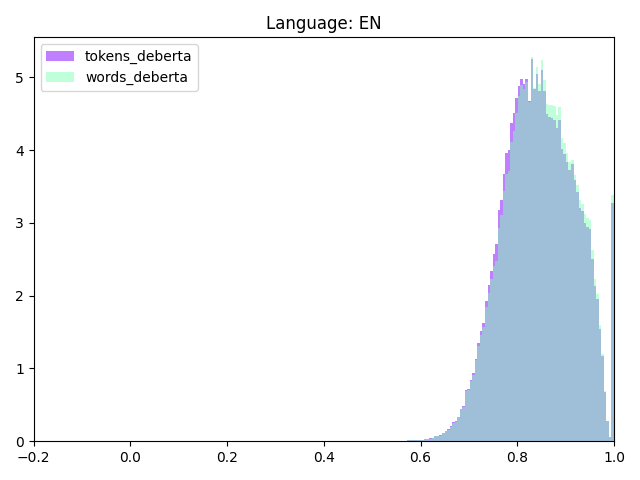}
    \end{minipage}
    &
    \begin{minipage}{0.47\linewidth}
    {\small Electra}
    
    \includegraphics[width=\linewidth,trim={0 0 0 1cm},clip]{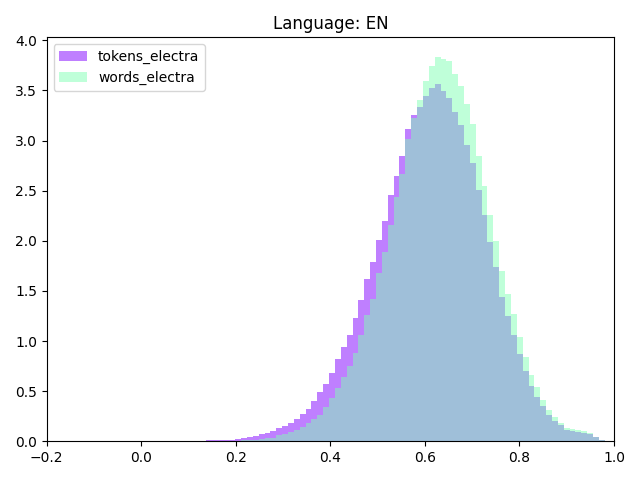}
    \end{minipage}
    \end{tabular}
    \vspace{-2mm}
    \caption{Histograms of cosine similarities for words and tokens in 1000 English sentences (results for French, German, Italian, Romanian and Spanish in appendix). The y-scales are different for each subplot for better visualization.}
    \label{fig:tokens_vs_words}
    \vspace{-2mm}
\end{figure}

For each sentence $s$ in the dataset, we compute the cosine similarity between the embeddings of every pair of tokens and every pair of words. The density histogram plots are shown in Figure \ref{fig:tokens_vs_words}. For BERT, the similarities among the token representations have a wider distribution, while they become tighter and centered on a higher mean for the optimized BERT variations, RoBERTa and DeBERTa, and for Electra. The word embeddings -- as averages of their token representations -- follow similar trends. These results show that with the changes in the models (relative to BERT), tokens and words within the same sentence become closer in the embedding space. According to the assumption that close points correspond to similar words and vice-versa, we must conclude that the optimizations over the BERT base model lead to more sharing of contextual information among the words in a sentence.

\begin{figure}[h]
    \begin{tabular}{cc}
    \begin{minipage}{0.47\linewidth}     
    {\small BERT}
    
    \includegraphics[width=\linewidth,trim={0 0 0 0.8cm},clip]{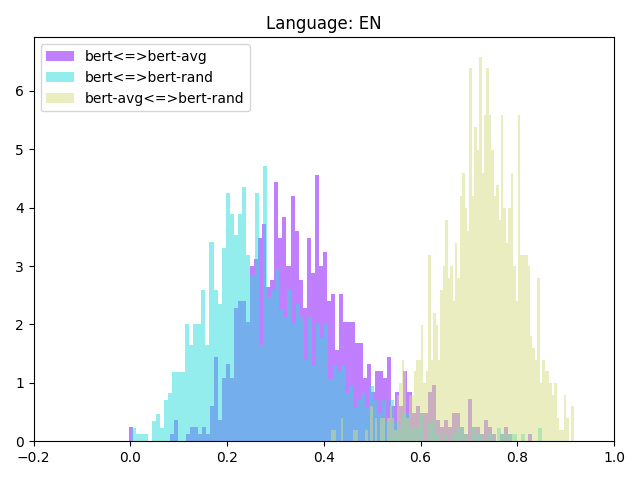}
    \end{minipage}
    &
    \begin{minipage}{0.47\linewidth}     
    {\small RoBERTa}
    
    \includegraphics[width=\linewidth,trim={0 0 0 1cm},clip]{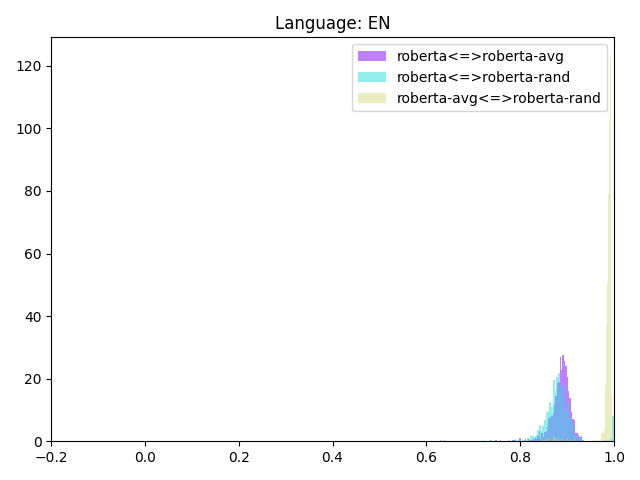}
    \end{minipage}
    
    \\ 
    
    \begin{minipage}{0.47\linewidth}
    {\small DeBERTa}
    
    \includegraphics[width=\linewidth,trim={0 0 0 0.8cm},clip]{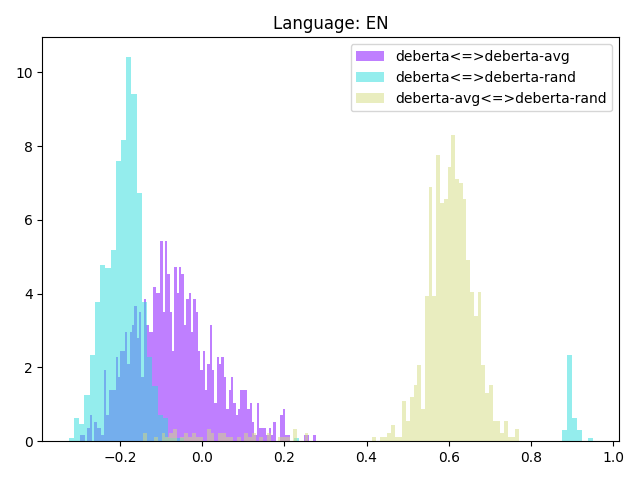}
    \end{minipage}
    &
    \begin{minipage}{0.47\linewidth}
    {\small Electra}
    
    \includegraphics[width=\linewidth,trim={0 0 0 1cm},clip]{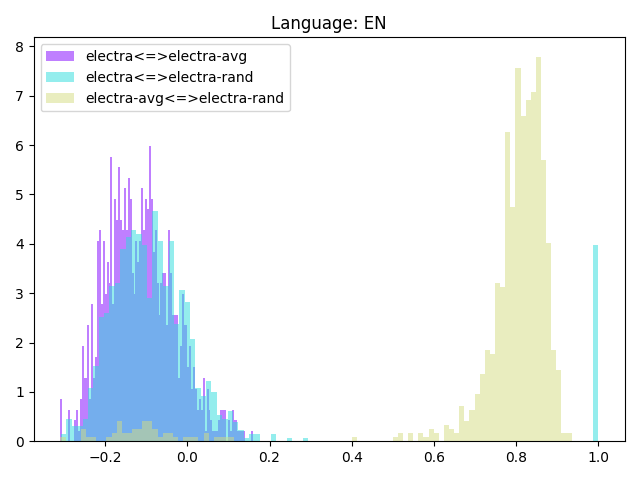}
    \end{minipage}
    \end{tabular}
    \vspace{-2mm}
    \caption{Histograms of cosine similarities computed for 1000 English sentences (results for French, German, Italian, Romanian and Spanish in appendix). In yellow are the distances between \savg~ and \trand, in blue the distances between \scls~ and \trand, and in purple the distances between \scls~ and \savg.
    The y-scales are different for each subplot for better visualization.
    \vspace{-5mm}
    }
    \label{fig:histograms}
\end{figure}

\vspace{-3mm}
\subsection{Distance between sentence representation variations in the embedding space}
\label{sec:sent_repr}

The next step is an analysis of the distance between the three types of embeddings ---token (\trand), averaged token (\savg), sentence (\scls)--- for several models from the BERT family. For each sentence $s$ in the dataset, we compute the cosine between every pair of representations \savg($s$), \scls($s$) and \trand($s$). Figure \ref{fig:histograms} shows the histograms of these comparisons.
This analysis also shows how the sentence representations change with the different training regimes and set-ups of the considered models.



For all models, \savg~ are very close to \trand, adding support to the observation from Section \ref{sec:tokens}, that tokens encode much contextual information. The holistic sentence embeddings \scls~ are quite dissimilar from both \savg~ and \trand~ for all models except for RoBERTa. The optimized training of RoBERTa has the effect of bringing all variations of the sentence embeddings closer together. 
%
It is interesting to note that \scls~ are almost orthogonal to \savg~ and \trand~ for both DeBERTa and Electra. Following the assumption of the smoothness of the embedding space, this may indicate that the holistic \scls~ embeddings encode different types of information than the contextual embeddings. We will test this in the next sections.


\section{Sentence representation comparisons on linguistics tasks}
\label{sec:holmes}

The previous analysis has shown that token embeddings encode much contextual information, and they, and the averaged token embeddings, are dissimilar from the embeddings of the special [CLS] token. This section shows experiments that investigate whether the similarities or differences noted in the embedding space analysis are reflected in their relative performances.

\subsection{Dataset and code}

We use the FlashHolmes benchmark \cite{waldis-etal-2024-holmes} to test the three embedding types. There are 216 tasks in morphology, syntax, semantics, discourse and reasoning, code to test new models, and a leaderboard for comparisons. The input of a task is obtained from the specified model, and the output is predicted using a classifier probe implemented as a linear NN layer. The results on these tasks will help determine what kind of information the three types of embedding encode.

\subsection{Performance comparisons}

\begin{figure*}[t]
    \includegraphics[width=\linewidth]{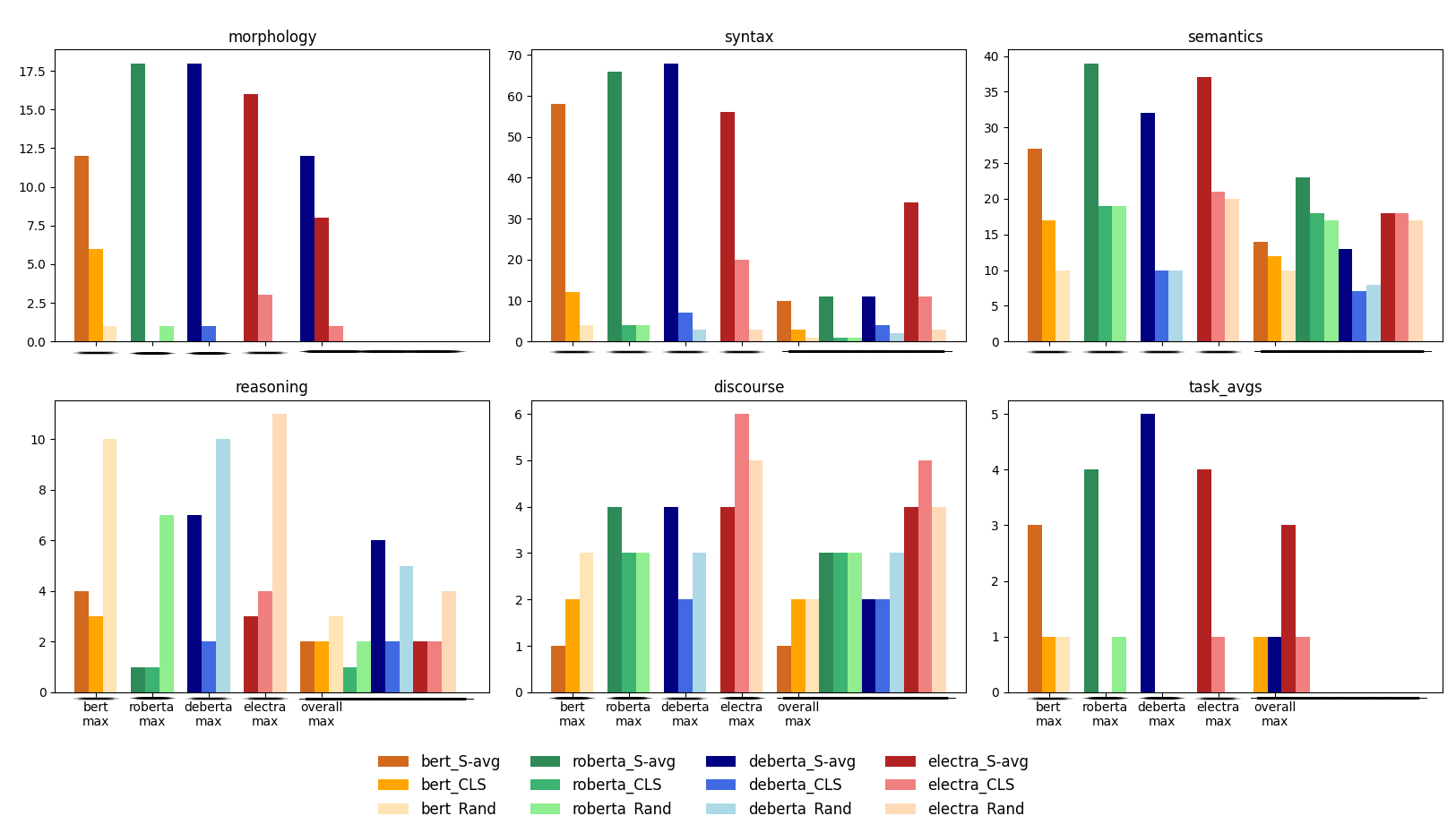}
    \caption{Statistics on the best sentence representation for each transformer, and overall for each task. The y-axis is the count of tasks for which a method performs best. For each transformer, we count the methods that performed best among the transformer's variations. If all variations have the same score, we count them only if they match the highest overall scores for the task.
    \vspace{-5mm}
    }
    \label{fig:task_method_stats}
\end{figure*}

Figure \ref{fig:task_method_stats} presents a summary of the performance of the different sentence representation methods for each task, and on the task averages.\footnote{For detailed (task-level) results see figures \ref{fig:holmes-fine-1}-\ref{fig:holmes-fine-2} and tables \ref{tab:morphology}--\ref{tab:reasoning} in the appendix.} The first four bar groups show the number of times each of the three sentence representation methods (represented as different shades of the model colour) has achieved the best performance for that model. The fifth group shows the number of times each sentence variation for each model has had the best overall performance. The sixth panel shows these statistics as averages over all tasks.

\begin{figure}[h]
    \centering
    \includegraphics[width=0.98\linewidth]{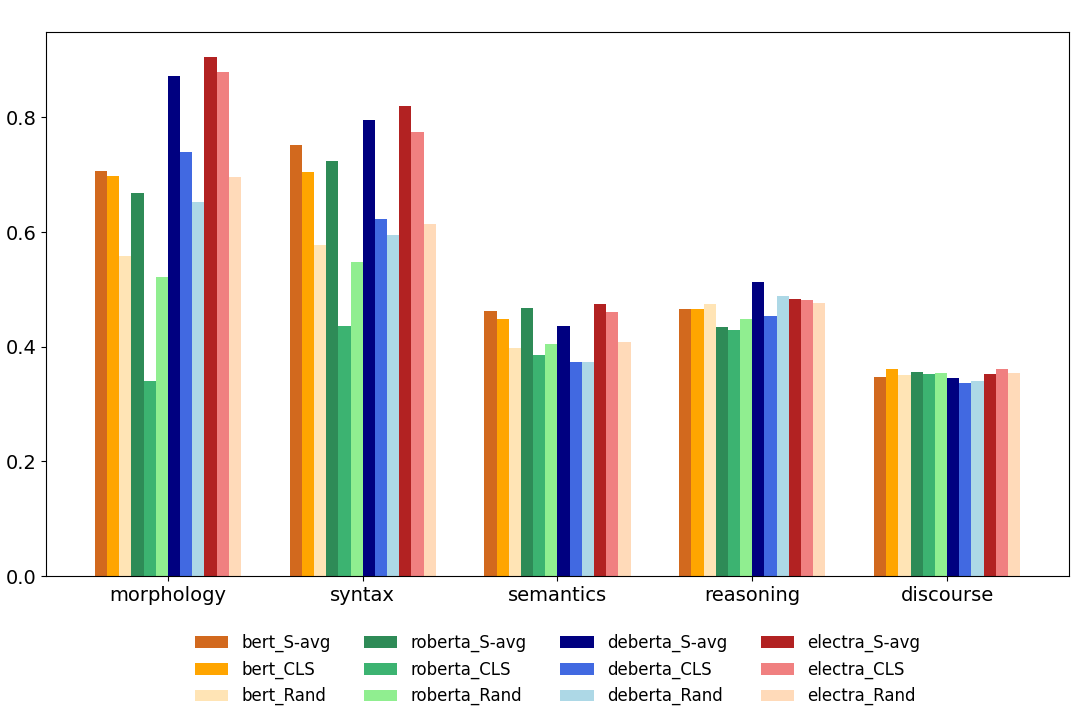}
    \caption{Comparison of embedding variations through average performance on the FlashHolmes benchmark}
    \label{fig:holmes-avg}
    \vspace{-3mm}
\end{figure}

On morphology, syntax and semantics the \savg~ has most frequently the highest performance for all models, while for reasoning and discourse \trand~ is often the best. However, the results in Figure \ref{fig:holmes-avg} show little variation in the results on the reasoning and discourse tasks, indicating that probably all results are close to the tasks baseline. For the morphology, syntax and semantics tasks, the performance of Electra's \savg~ and \scls~ sentence representations are very close in terms of average performance, while \trand is much lower. This seemingly contradicts the analysis in Section \ref{sec:sent_repr} which shows that \savg~ and \scls~ representations are almost orthogonal, while \savg~ and \trand~ are very close. This supports the hypothesis that the geometry of the embedding space is not informative about the linguistic properties encoded by the embeddings. There can be an alternative explanation for the close performance of \savg~ and \scls: they encode the same information, but in different ways. We explore this in the next section.



\begin{figure*}[t]
\vspace{-1cm}
\includegraphics[width=\linewidth]{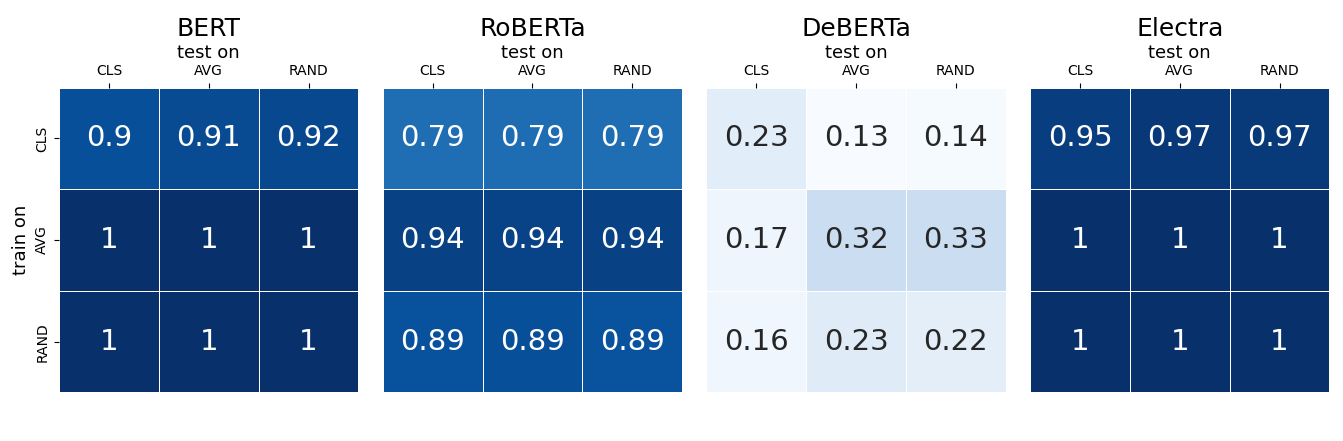}
\vspace{-1cm}
\caption{Comparison between models using \savg, \scls~ and \trand~ in detecting the sentence chunk structure in terms of average F1 scores over three runs. Detailed results in table \ref{tab:sent_str} in the appendix.}
\label{fig:structure_Res}
\end{figure*}

\savg~ and \scls~ embeddings have high and close performance for most task types.
Compared with the analysis of the embeddings as vectors in the embedding space, this result is unexpected, as for Electra and DeBERTa in particular, the \scls~ and the \savg~ embeddings are almost orthogonal. Not only these embeddings have similar performance, but even for variations of the same task \savg~ gives best results for one task, and \scls~ gives best results for the other. For RoBERTa, where the cosine similarity between these two variations is very high, their relative performance is very different\footnote{e.g. blimp\_determiner\_noun\_agreement\_with\_adj\_irre gular\_(1 and 2), blimp\_irregular\_plural\_subject\_verb\_agree ment\_(1 and 2), blimp\_principle\_A\_case\_(1 and 2), blimp\_principle\_A\_domain\_(1 and 2)}.

\section{Probing for structure}
\label{sec:structure}

The previous experiments on a variety of morphological, syntactic, semantic, discourse and reasoning tasks within the FlashHolmes benchmark 
show very close performance on the \savg~ and \scls~ variations. In light of the analysis of the relative position of embeddings in the embedding space, these results are surprising: for Electra and DeBERTa in particular, the two representations seem to be almost orthogonal (see Figure \ref{fig:histograms}). An explanation could be that the same type of clues necessary to solve these tasks is encoded in different manners in the two types of representations. We test whether this is indeed the case, by focusing on sentence structure. Sentence structure is complex, relying on clues about phrase boundaries and phrase properties. We test whether information about sentence structure can be detected in the three sentence representation variations, and whether it is encoded in a similar manner. For this we use the approach of \citet{nastase-merlo-2024-identifiable}, who have shown that some types of structural information -- noun, prepositional, or verb phrase (chunks) structure -- is recoverable from sentence representations. We use their code and data to investigate the sentence representations \footnote{\url{https://github.com/CLCL-Geneva/BLM-SNFDisentangling}}. Experiments that use the training and test data encoded using the same representation type will reveal whether the targeted sentence structure is identifiable. Cross-representation experiments -- using the training data encoded with one type of representation, and the test data encoded with the other two types -- will show whether the necessary information do detect structure is encoded in the same way.

\begin{figure*}[t]
    \begin{tabular}{cc}
    \begin{minipage}{0.47\linewidth}     
    {\small BERT}

    \includegraphics[width=\linewidth,trim={1.5cm 3cm 1cm 1.2cm},clip]{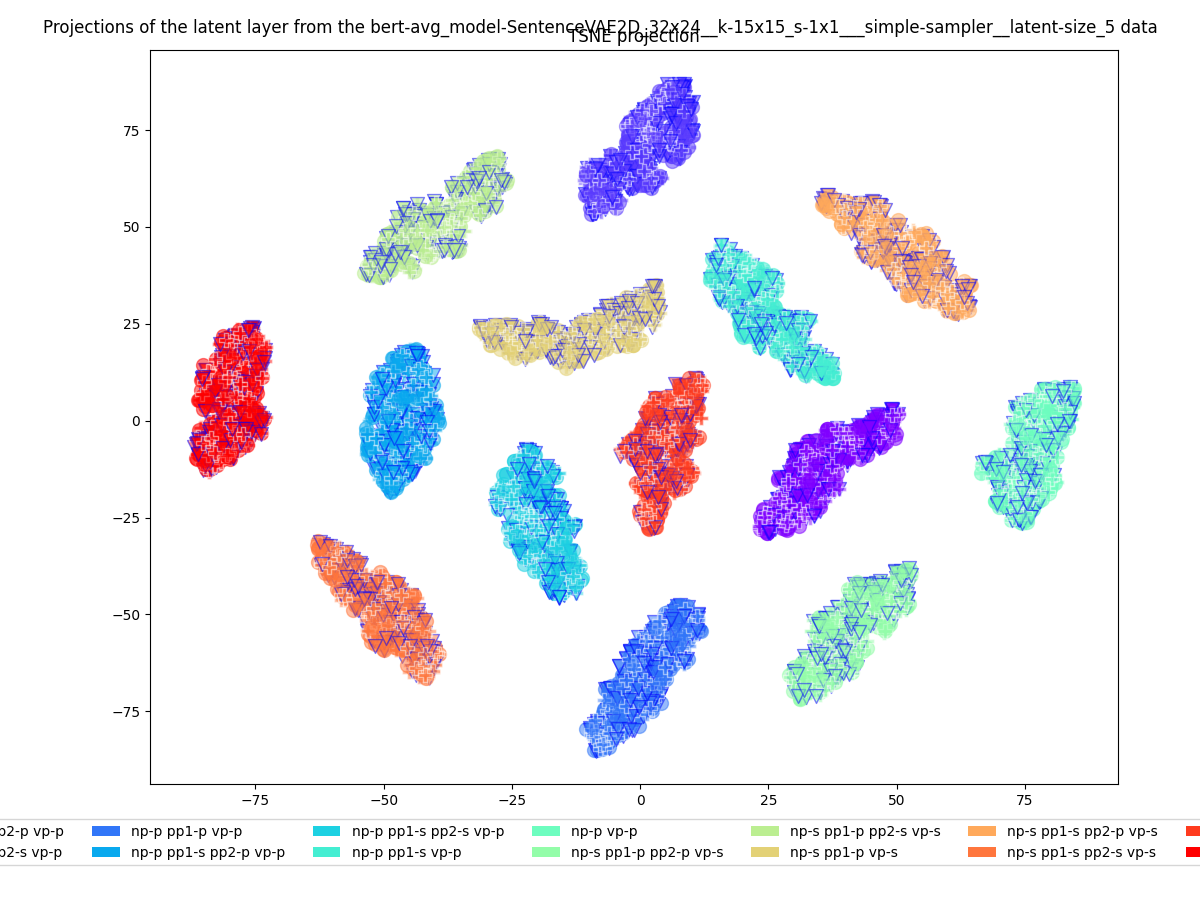}
    \end{minipage}
    &
    \begin{minipage}{0.47\linewidth}     
    {\small RoBERTa} 

    \includegraphics[width=\linewidth,trim={0 3cm 0 1.5cm},clip]{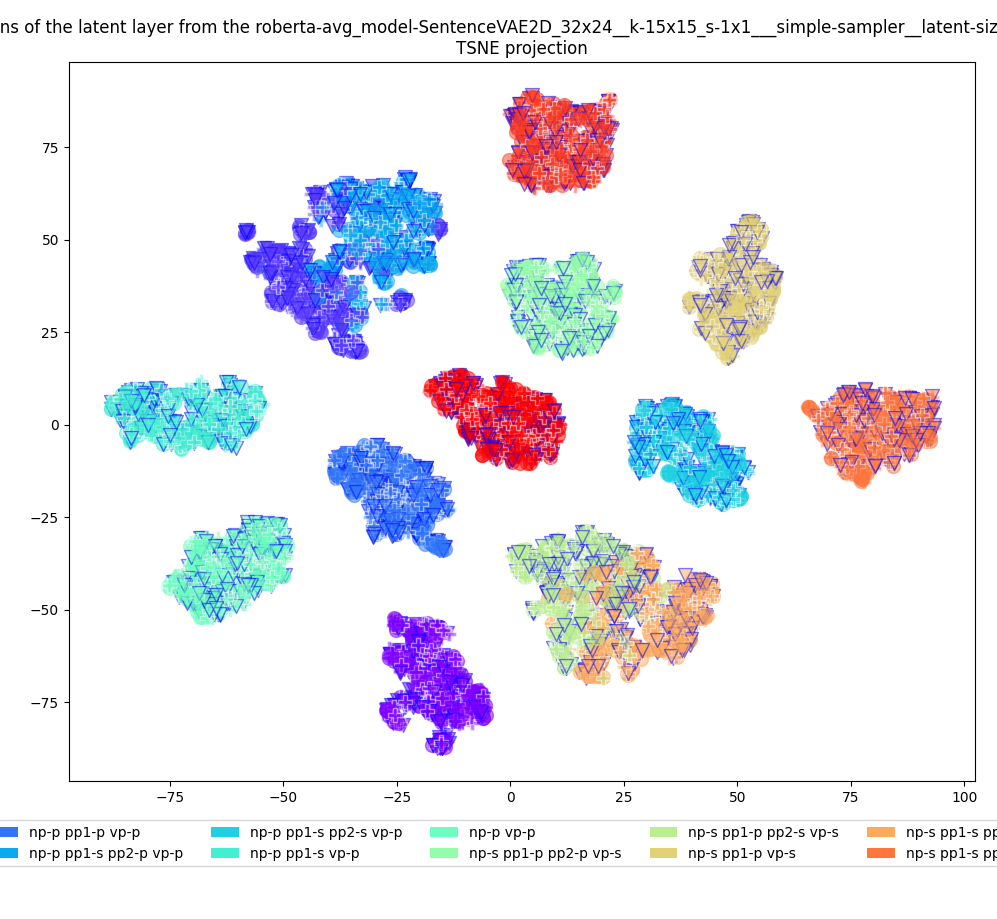}
    \end{minipage}
    
    \\ 
    
    \begin{minipage}{0.47\linewidth}
    {\small DeBERTa} 

    \includegraphics[width=\linewidth,trim={0 2.7cm 0 1.5cm},clip]{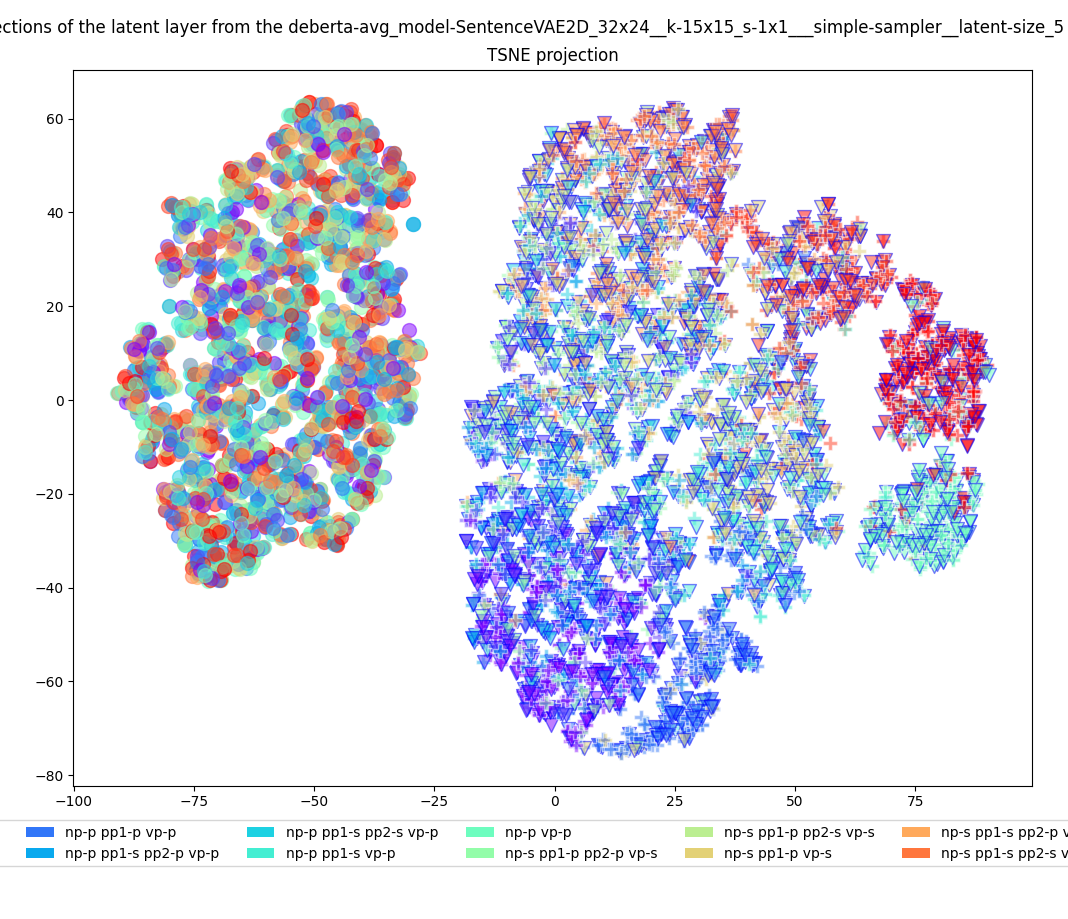}
    \end{minipage}
    &
    \begin{minipage}{0.47\linewidth}
    {\small Electra} 

    \includegraphics[width=\linewidth,trim={0 3.2cm 1cm 1.5cm},clip]{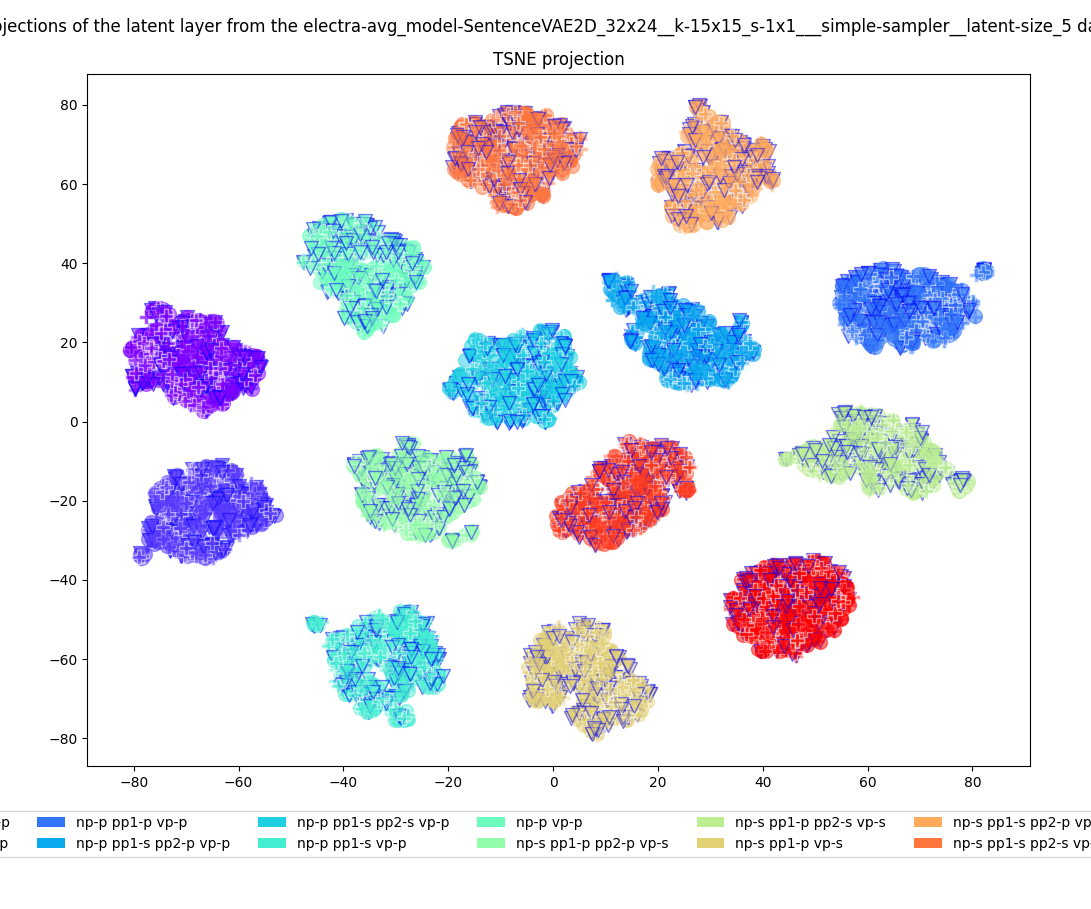}
    \end{minipage}
    \end{tabular}
    \includegraphics[width=\linewidth]{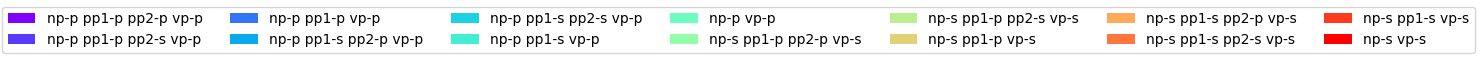}
    \caption{Comparison between models using \savg~ ($\bigcirc$), \scls~ ($\bigtriangledown$) and \trand~ ($+$) in detecting the sentence chunk structure. tSNE plots of the latent layer vectors of the training data represented using \savg~, \scls~ and \trand, obtained from a model trained on the \savg~ representation. The latent layer vectors are expected to encode the targeted information, i.e. the chunk structure. We note very sharp clusters for BERT and Electra}
    \label{fig:structure_plots}
\end{figure*}

\subsection{Dataset and code}

\paragraph{The dataset} consists of English sentences with the syntactic pattern \texttt{np (pp$_1$ (pp$_2$)) vp}\footnote{The pattern uses the BNF notation: pp$_1$ and pp$_2$ may be included or not, pp$_2$ may be included only if pp1 is included}, where each \texttt{np, pp$_1$, pp$_2$, vp} can be in the singular or plural form, and the subject (np) always agrees with the verb (vp). There are 4004 sentences evenly split across the 14 chunk patterns. An instance is built for each sentence as a triple $(in, out^+, Out^-)$, where $in$ is an input sentence with chunk structure $p$, $out^+$ is a sentence different than $in$ but with the same chunk structure $p$, and $Out^-$ is a set of $N_{negs}=7$ sentences each of which has a chunk pattern different from $p$ (and different from each other). The 4004 instances are split into train:dev:test -- 2576:630:798.

\paragraph{The system} is a variational encoder-decoder. The encoder consists of a CNN layer that splits the input sentence embedding into layers of information, which it then compresses using a linear layer into a small latent representation. The decoder is a mirror image of the encoder, but unlike a regular variational auto-encoder, it does not reconstruct the input. Sentence embeddings have 768 dimensions, and are compressed on the latent layer to size 5. To encourage the sentence chunk structure to be encoded in the latent layer, each input $in$ will be decoded into $out^+$ -- a different sentence but with the same chunk structure -- using the $N_{negs}$ sentences with different structure than the input as contrastive examples. While the system receives a supervision signal -- the correct output -- it does not receive explicit information about a sentence's structure. While there are 14 structure patterns in the data, each instance contains 7 randomly chosen negative instances. So with respect to the sentence structure, there is only indirect supervision.

\subsection{Performance comparison}

We apply this approach to the provided sentence data when using the \scls, \savg~ and \trand~ sentence representations. We present two perspectives of the performance: (i) in terms of F1 averages over three runs (how well does the system perform in building a sentence representation closest to the correct one) shown in Figure \ref{fig:structure_Res} and (ii) an analysis of the latent layer of the system, shown in Figure~\ref{fig:structure_plots}.

Despite high results on the syntactic and semantic Holmes tasks, detecting the chunk structure is not successful on the DeBERTa embeddings. This may be because of DeBERTa's optimized training, with disentangled attention matrices and token embeddings with separate position and content sections, which leads to differently organized embeddings than BERT, RoBERTa and Electra. BERT and Electra in particular show very high results, with results on \trand~ even higher than \scls. 

For the purpose of determining whether the variations in sentence representation encode the same information in the same manner, we look at the cross-testing results -- training on one representation, and testing on the others (Figure \ref{fig:structure_Res}). Despite the differences revealed by the cosine similarity analysis, where for Electra the \scls~ representations are almost orthogonal to \savg~ and \trand, these experiments show that all three representations encode information about the chunk pattern in a sentence, and moreover, this information is encoded in the same manner. Additional support for this hypothesis comes from the analysis of the latent layer. Figure \ref{fig:structure_plots} show the tSNE projection\footnote{We chose the tSNE projection because it preserves the neighbourhoods. We use the tSNE implementation from scikit-learn, with 2 components and default parameters.} of the latent representations obtained from a model trained using the \savg~ sentence representations. After training on the training data represented with \savg, we pass through the encoder the sentences in the dataset encoded with all sentence representation variations, collect all latents and project them in 2D using tSNE. 

The results on the task and the analysis of the latent vectors provide complementary views about whether structural information is encoded in the sentence embeddings. We may obtain high results on the task (choosing the sentence with the same structure) while for each sentence representation the vectors are mapped into a different area in the latent space. The tSNE plots show that in fact the different types of sentence representations are mapped onto shared regions. This is highly significant: it indicates that the clues based on which the structure is detected is encoded in a similar manner in all the three sentence representations, regardless of the differences among them we have noted during the previous experiments. 

\citet{Nastase_CLIC-IT2024-2_2024} have shown, through experiments on several languages, that sentence embeddings do not encode chunk structure as an abstraction, but rather linguistic clues -- such as phrase boundaries and number information -- that can be assembled into the chunk structure. Considering this, and the results in Figure \ref{fig:structure_Res} and the plots in Figure \ref{fig:structure_plots}, this indicates that the \savg~ and \scls~  encode the information about phrase boundaries and number in the same manner and in the same location for BERT and Electra in particular.


\section{Word and text representations in the embedding space}


\paragraph{The evolution of the embedding space} 
Procedurally and scale-wise, we have come a long way from the first distributional models of language inspired by \newcite{Harris54} and \newcite{Firth57}, with tens of thousands of symbolic dimensions computed over a small (relative to what is used today) corpus \cite{schuetze1992}.
Symbolic dimensions are interpretable, but also brittle, and overlapping, and were tackled using clustering \cite{pantel02,Blei2003}, or dimensionality reduction \cite{Furnas88,landauer1997solution,Jolliffe02,Blei2003}. \newcite{landauer1997solution}'s approach can be viewed as 3 layer neural network, but \newcite{Bengio2003} first used a neural network to encode the probability function of word sequences in terms of the feature vectors of the words in the sequence. 
Pre-trained word embeddings, started with \cite{Mikolov2013b,Mikolov2013efficient}, have been shown to encode syntactic and semantic information, as regularities in the relative position of words in the low-dimensional vector space 
\cite{Ethayarajh2019}. Currently, contextual embeddings are obtained with transformer-based models \cite{vaswani2017attention}. Models from the BERT family 
\cite{devlin-etal-2019-bert} produce token embeddings and sentence representations as the embedding of a special [CLS] token. Generative language models do not produce sentence embeddings as such, although approximations can be obtained using word definition-like prompts \cite{jiang-etal-2024-scaling}.

Embedding dimensions encode some linguistic information: shallow information about sentences \citet{nikolaev-pado-2023-universe}, sentence-level information \citep{tenney-etal-2019-learn-from-context}, including syntactic structure -- reflected as relative positions in the embedding space that parallel a syntactic tree \citep{hewitt-manning-2019-structural,chi-etal-2020-finding}. Deeper exploration through simple classification probes (consisting of one linear NN), has shown that predicate embeddings contain information about their semantic roles structure \citep{conia-navigli-2022-probing,carvalho-etal-2023-learning}, embeddings of nouns encode subjecthood and objecthood \citep{papadimitriou-etal-2021-deep}, and syntactic and semantic information can be teased apart \cite{mercatali-freitas-2021-disentangling-generative,bao-etal-2019-generating,chen-etal-2019-multi}. 

\paragraph{The geometry of the embedding space}



The embedding space of tokens appears to be anisotropic \cite{mimno-thompson-2017-strange,timkey-van-schijndel-2021-bark,cai-etal-2021-isotropy}, which can adversely influence model training and fine-tuning. Anisotropy could be caused by a few dominant dimensions, that can skew the similarity profile of the space \cite{timkey-van-schijndel-2021-bark}. However, the embedding space actually contains isotropic clusters and lower-dimensional manifolds that reflect word frequency properties \cite{cai-etal-2021-isotropy}.

\section{Discussion and Conclusions}

The output of pretrained language models provide embeddings for individual tokens, and a holistic sentence embedding as the embedding of a special token. A sentence is often represented through the averaged embeddings of its tokens, or through this special token embedding. In the extreme, we could even use the embedding of a random token to represent the sentence. In this work, we explored how different, or similar, these three types of representations are, and what kind of information they encode. What we found is a complex picture. Shallow analysis through cosine similarity measures shows how distinct these three representations are, and how they change relative to each other from a baseline system (BERT) with various optimizations (RoBERTa), internal organization changes (DeBERTa) or changes in the training regimen (Electra) of the system. These shallow differences or similarities are not reflected in benchmarks on five types of NLP tasks, where seemingly orthogonal representations lead to very similar results on many tasks. 

The close performance of the seemingly very distinct sentence representations raises another question: do they encode similar information in a similar manner, or the results come from exploiting different, or differently encoded, cues? Experiments in detecting a sentence's chunk structure -- the sequence of NP/VP/PP phrases and their grammatical number attributes -- showed that in fact information relevant for reconstructing this structure is encoded in the same manner, as a system trained on one sentence representation has a very similar performance when tested on the other.

The experiments presented in this paper add to the complex picture of what kind of information the embeddings induced by pretrained transformer models encode, and how. The results show that embedding dimensions do not encode linguistic information superficially, rather linguistic features are encoded through more complex weighted combinations of features. Some of these are shared among all tokens in a sentence, and within the holistic sentence embedding.

\section{Limitations}

\paragraph{Synthetic data with 14 structure patterns}
To study the deeper question of whether the different sentence embedding variations encode sentence structure the same way, we have used a synthetic dataset, with limited variation in sentence structure, expressed as a sequence of chunks, or phrases. In future work we plan to investigate what level of structure complexity can be recovered from these embeddings, and whether at some complexity level, differences among the embedding variations becomes apparent.

\paragraph{Raw output of transformer models}
We have focused on four pretrained models from the BERT family, and analyzed their sentence embedding space through cosine similarity, solving tasks and detecting sentence structure. We have excluded from the related work and analysis sentence transformers, which fine-tune sentence embeddings for similarity. Our aim was to study the raw output of the transformer models, and understand the properties of the different types of embeddings they induce.

\paragraph{No generative language models} We focused on models form the BERT family because they explicitly induce sentence representations as the embedding of a special token, or they can be computed as averaged token embeddings. It was crucial for our experiments to have several sentence representation variations to compare. Generative models do not produce sentence embeddings. Representations approximating such representations have been induced using word definition-like prompts \cite{jiang-etal-2024-scaling,zhang2024simple}. Our interest has been to study more fundamental properties of transformer-based models, rather than test the performance of sentence representation approximations.

\paragraph{Cosine similarity} We reported analyses in terms of cosine similarity which is the most commonly used in the training objective. The analysis in terms of euclidean distance did not provide additional insights, so it was not included.

\bibliographystyle{natbib}
\bibliography{custom,SRBN2,anthology}

\appendix

\onecolumn

\section{Words vs. token embedding similarities distribution}

\begin{figure}[h]
    \begin{tabular}{lcccc} \\
    Lang & BERT & RoBERTa & DeBERTa & Electra \\ \hline
    \vcent{English}   & \includegraphics[width=0.2\linewidth,trim={0 0 0 1cm},clip]{figs/token_embeddings_bert_EN.png} & \includegraphics[width=0.2\linewidth,trim={0 0 0 1cm},clip]{figs/token_embeddings_roberta_EN.png} & \includegraphics[width=0.2\linewidth,trim={0 0 0 1cm},clip]{figs/token_embeddings_deberta_EN.png} & \includegraphics[width=0.2\linewidth,trim={0 0 0 1cm},clip]{figs/token_embeddings_electra_EN.png} \\ \hline
    \vcent{French}    & \includegraphics[width=0.2\linewidth,trim={0 0 0 1cm},clip]{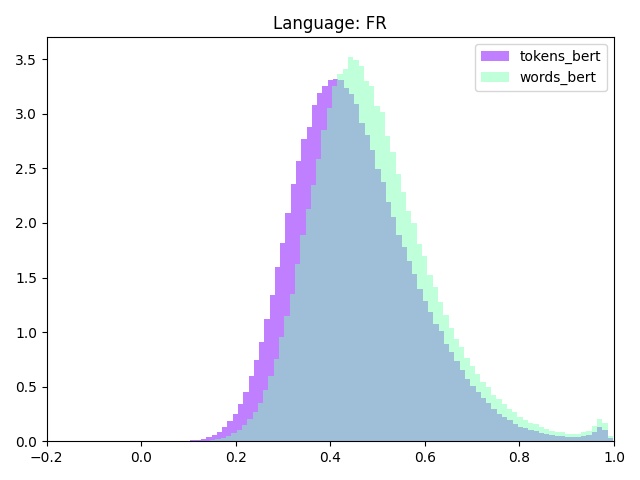} & \includegraphics[width=0.2\linewidth,trim={0 0 0 1cm},clip]{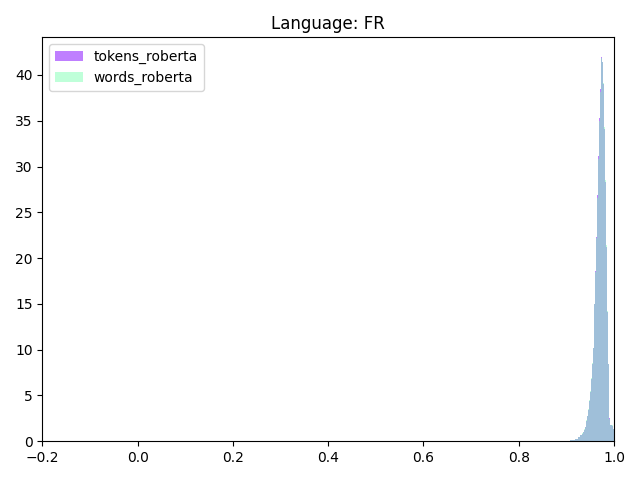} & \includegraphics[width=0.2\linewidth,trim={0 0 0 1cm},clip]{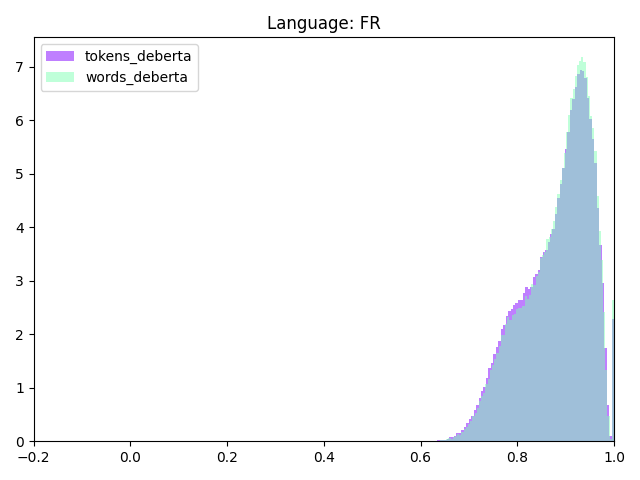} & \includegraphics[width=0.2\linewidth,trim={0 0 0 1cm},clip]{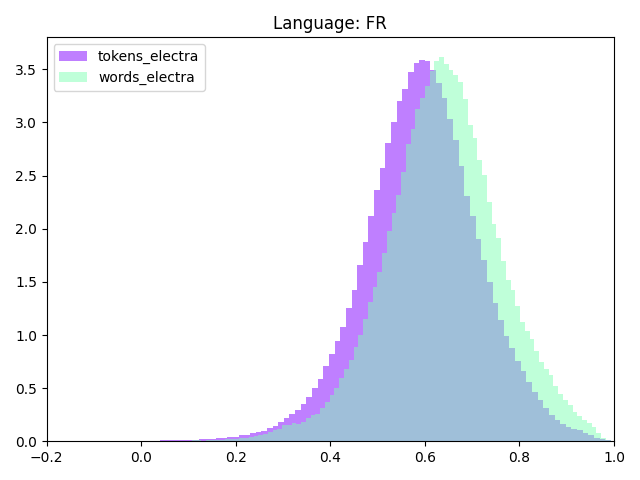} \\ \hline
    \vcent{German}   & \includegraphics[width=0.2\linewidth,trim={0 0 0 1cm},clip]{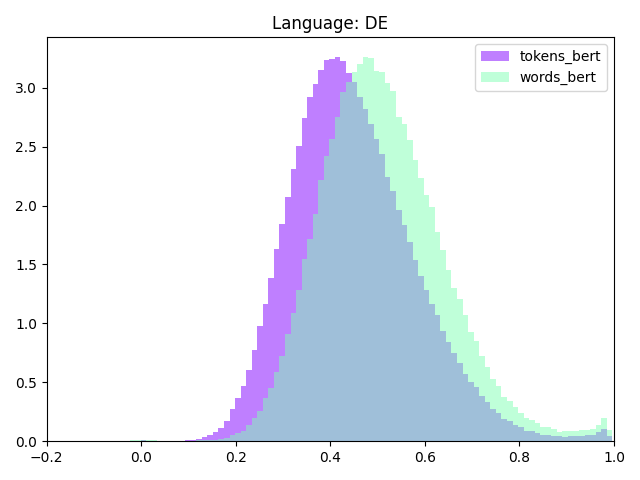} & \includegraphics[width=0.2\linewidth,trim={0 0 0 1cm},clip]{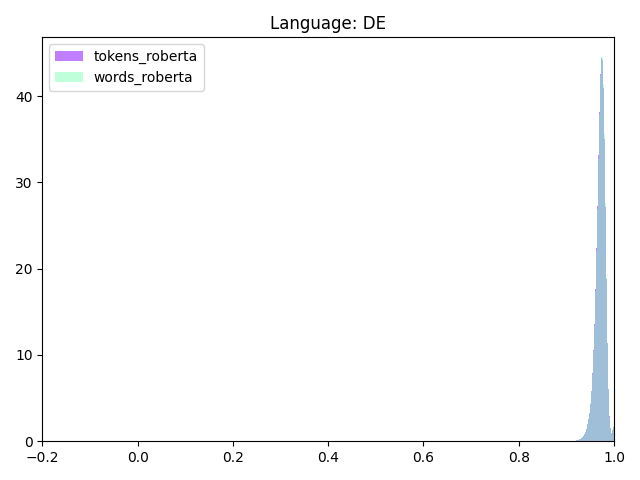} & \includegraphics[width=0.2\linewidth,trim={0 0 0 1cm},clip]{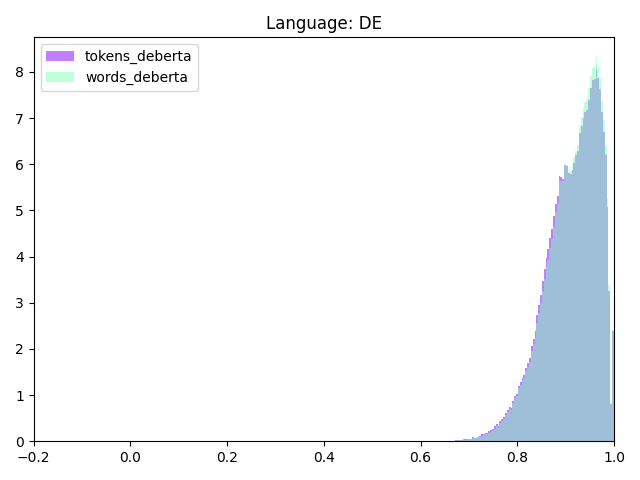} & \includegraphics[width=0.2\linewidth,trim={0 0 0 1cm},clip]{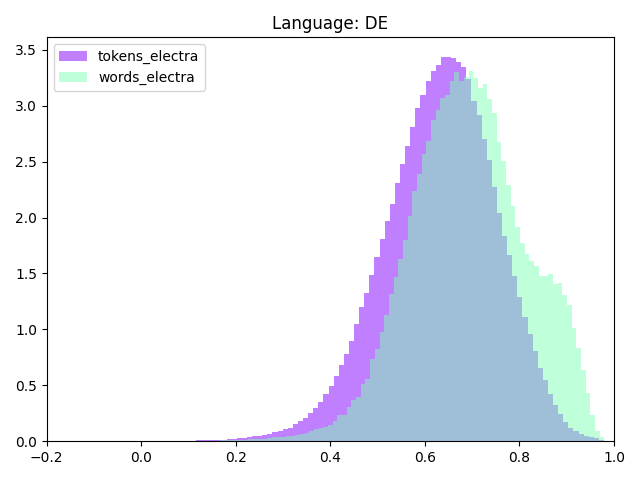} \\ \hline
    \vcent{Italian}   & \includegraphics[width=0.2\linewidth,trim={0 0 0 1cm},clip]{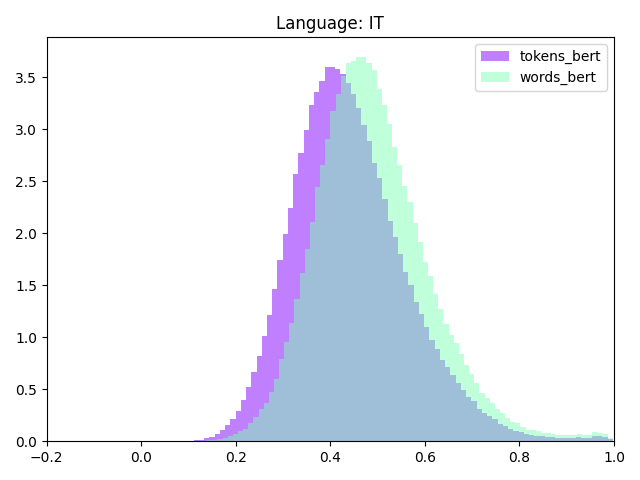} & \includegraphics[width=0.2\linewidth,trim={0 0 0 1cm},clip]{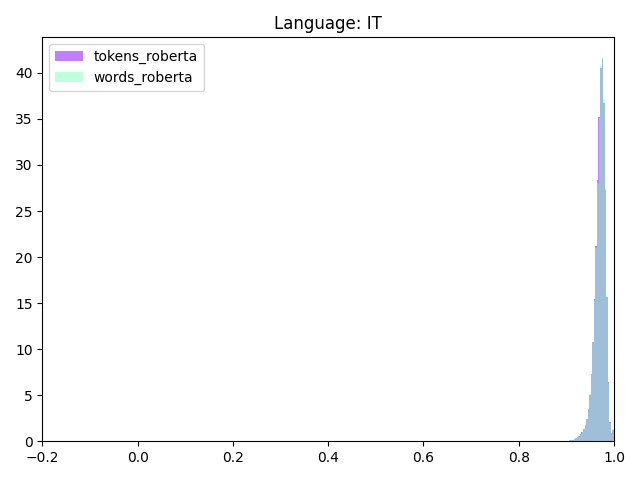} & \includegraphics[width=0.2\linewidth,trim={0 0 0 1cm},clip]{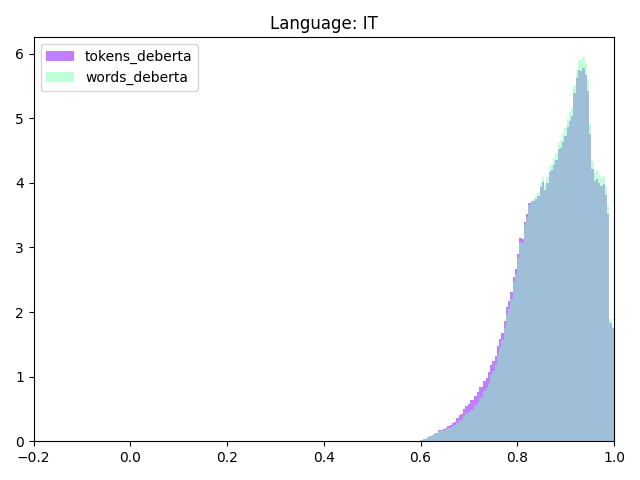} & \includegraphics[width=0.2\linewidth,trim={0 0 0 1cm},clip]{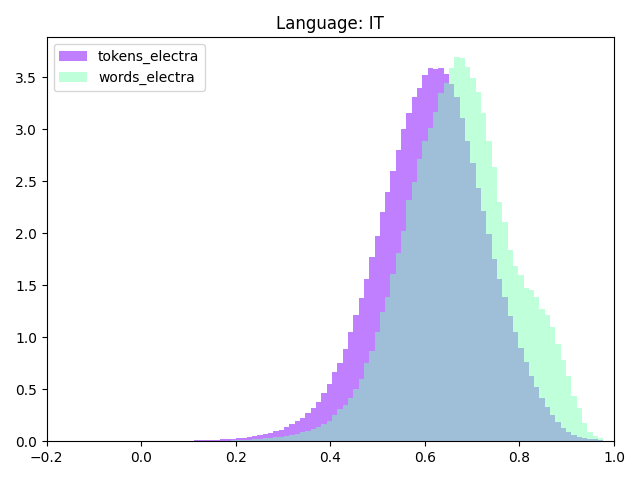} \\ \hline
    \vcent{Romanian}   & \includegraphics[width=0.2\linewidth,trim={0 0 0 1cm},clip]{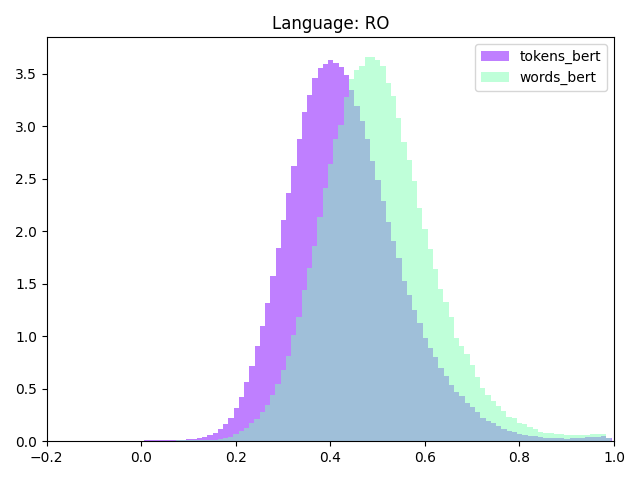} & \includegraphics[width=0.2\linewidth,trim={0 0 0 1cm},clip]{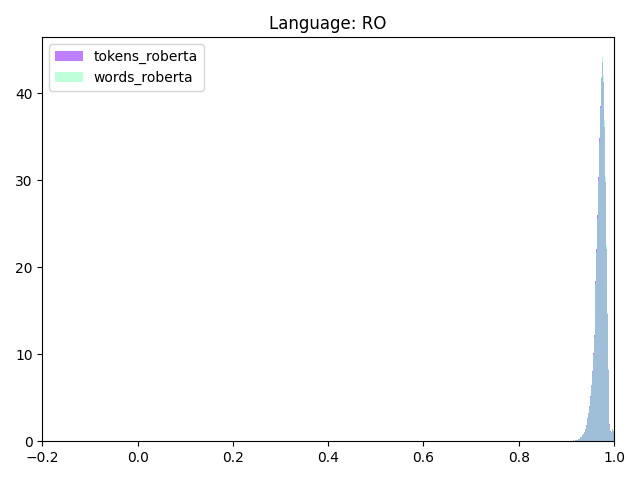} & \includegraphics[width=0.2\linewidth,trim={0 0 0 1cm},clip]{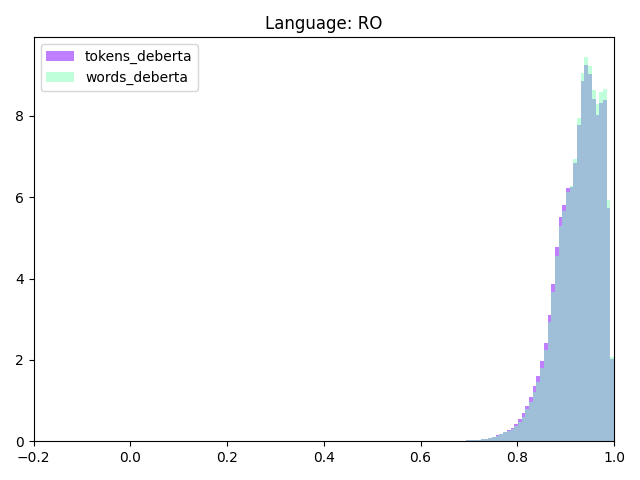} & \includegraphics[width=0.2\linewidth,trim={0 0 0 1cm},clip]{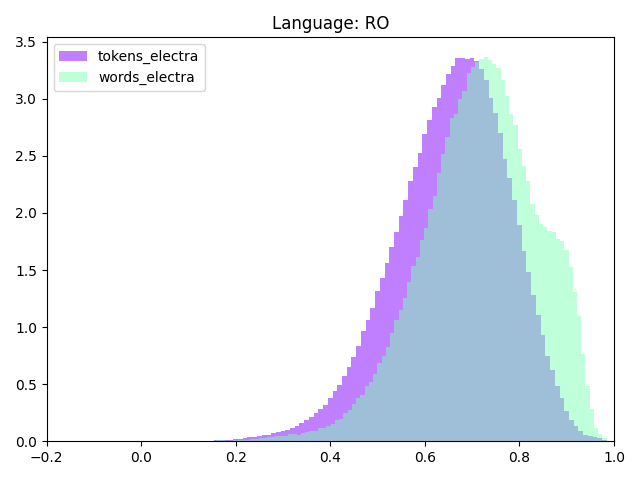} \\ \hline
    \vcent{Spanish}   & \includegraphics[width=0.2\linewidth,trim={0 0 0 1cm},clip]{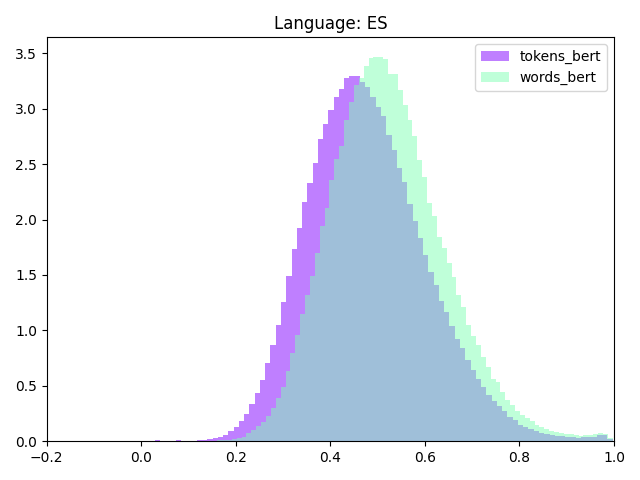} & \includegraphics[width=0.2\linewidth,trim={0 0 0 1cm},clip]{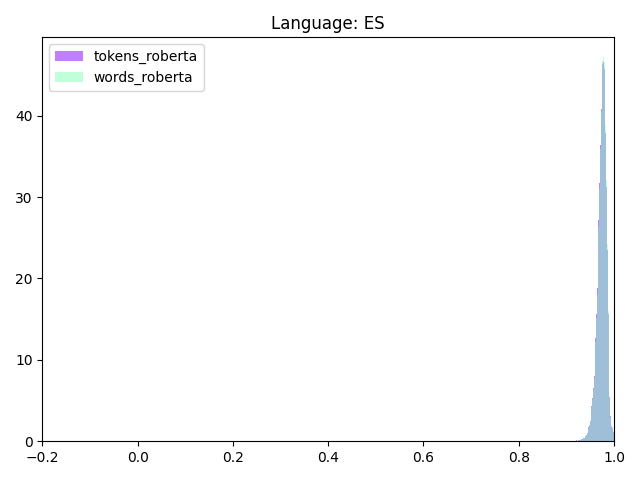} & \includegraphics[width=0.2\linewidth,trim={0 0 0 1cm},clip]{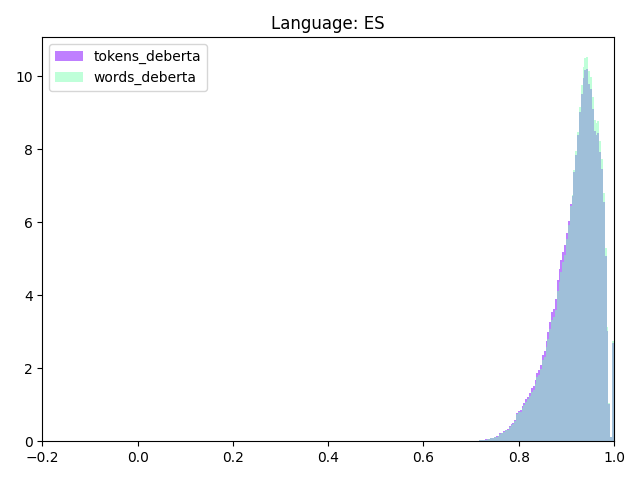} & \includegraphics[width=0.2\linewidth,trim={0 0 0 1cm},clip]{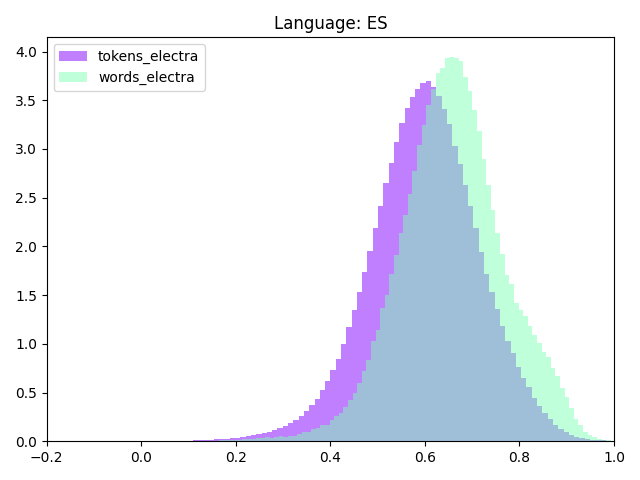} \\ \hline
    \end{tabular}
    \caption{Cosine similarities histograms computed for words and tokens from 1000 English/French/German/Italian/Romanian/Spanish.}
    \label{fig:words_vs_tokens_nlang}
\end{figure}

Figure \ref{fig:words_vs_tokens_nlang} shows a comparison between the distribution of token and word similarities within the same sentence. A tighter distribution -- as displayed by RoBERTa embeddings -- indicates that all contextual embeddings are closer to each other, and thus encode more contextual information. BERT and Electra embeddings display distributions with larger standard deviation, indicating that there is more variation in the information encoded in the individual tokens and words. Electra token/word distances have a higher mean, indicating that these embeddings encode more contextual information than BERT ones. All distributions have a high spike close to 0. These pairs include punctuation and "suffix" tokens.

\newpage 

\section{Multilingual embeddings comparison}

\begin{figure}[h]
    \begin{tabular}{lcccc} \\
    Lang & BERT & RoBERTa & DeBERTa & Electra \\ \hline
    \vcent{English}   & \includegraphics[width=0.2\linewidth,trim={0 0 0 1cm},clip]{figs/sentence_embeddings_bert_EN.png} & \includegraphics[width=0.2\linewidth,trim={0 0 0 1cm},clip]{figs/sentence_embeddings_roberta_EN.png} & \includegraphics[width=0.2\linewidth,trim={0 0 0 1cm},clip]{figs/sentence_embeddings_deberta_EN.png} & \includegraphics[width=0.2\linewidth,trim={0 0 0 1cm},clip]{figs/sentence_embeddings_electra_EN.png} \\ \hline
    \vcent{French}   & \includegraphics[width=0.2\linewidth,trim={0 0 0 1cm},clip]{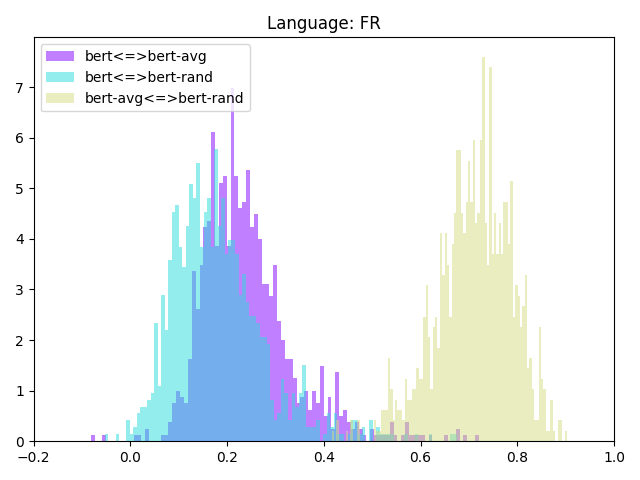} & \includegraphics[width=0.2\linewidth,trim={0 0 0 1cm},clip]{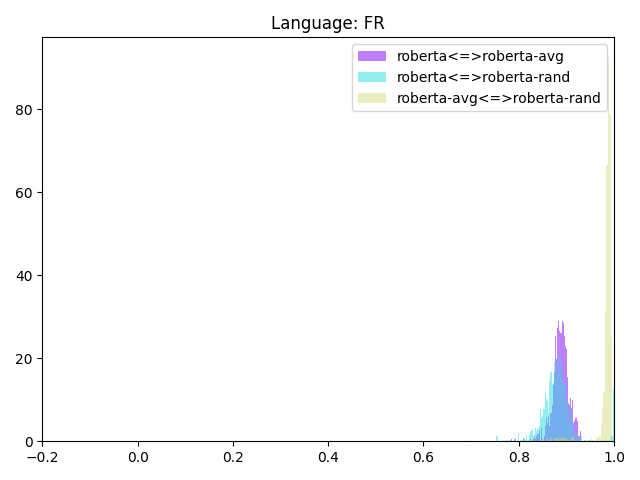} & \includegraphics[width=0.2\linewidth,trim={0 0 0 1cm},clip]{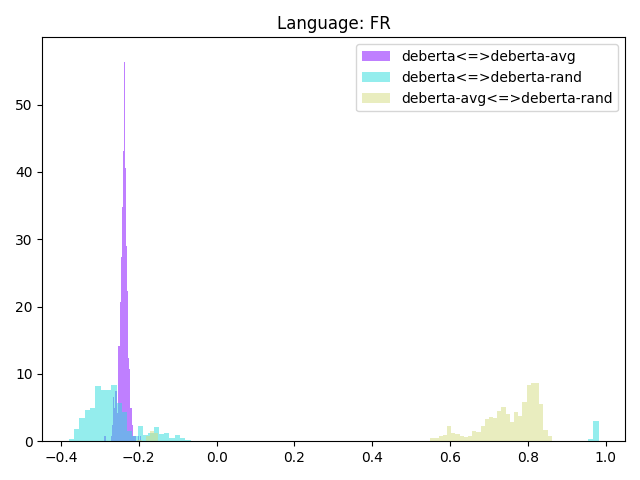} & \includegraphics[width=0.2\linewidth,trim={0 0 0 1cm},clip]{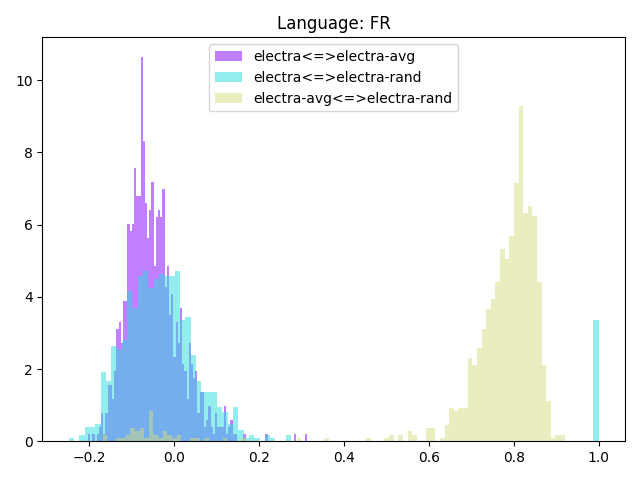} \\ \hline
    \vcent{German}   & \includegraphics[width=0.2\linewidth,trim={0 0 0 1cm},clip]{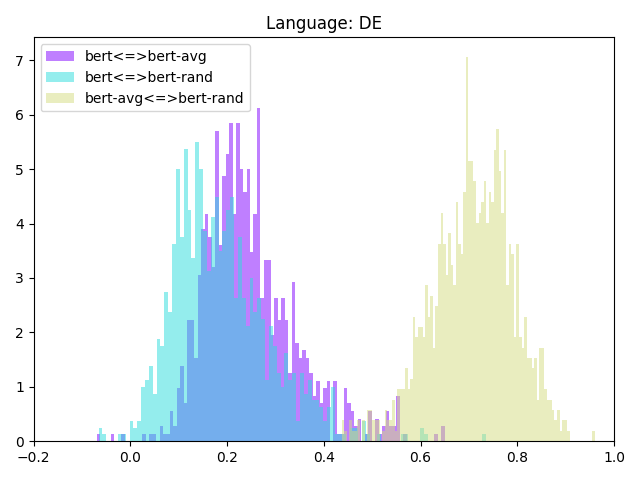} & \includegraphics[width=0.2\linewidth,trim={0 0 0 1cm},clip]{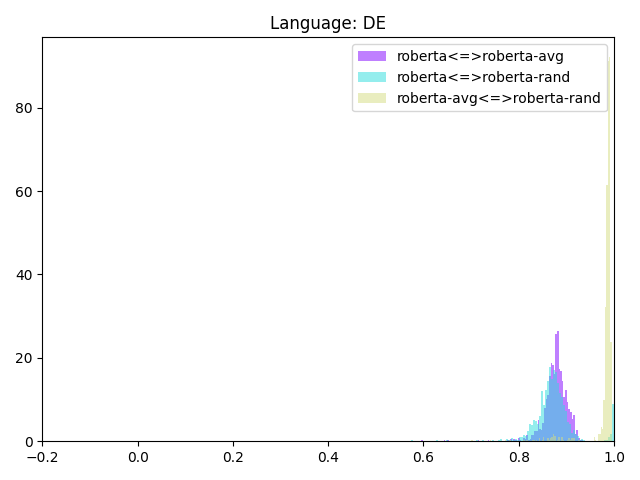} & \includegraphics[width=0.2\linewidth,trim={0 0 0 1cm},clip]{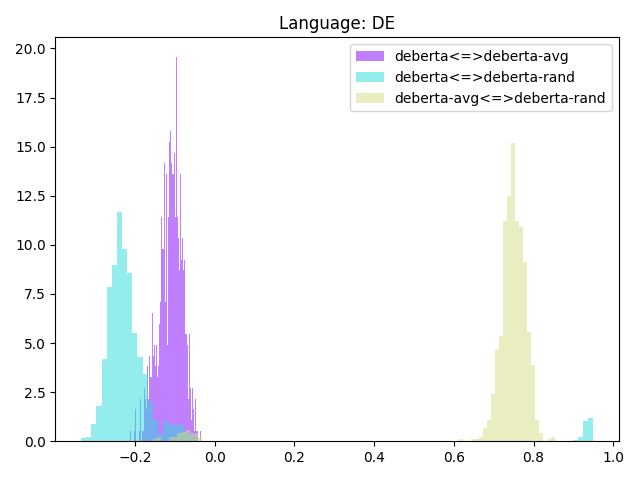} & \includegraphics[width=0.2\linewidth,trim={0 0 0 1cm},clip]{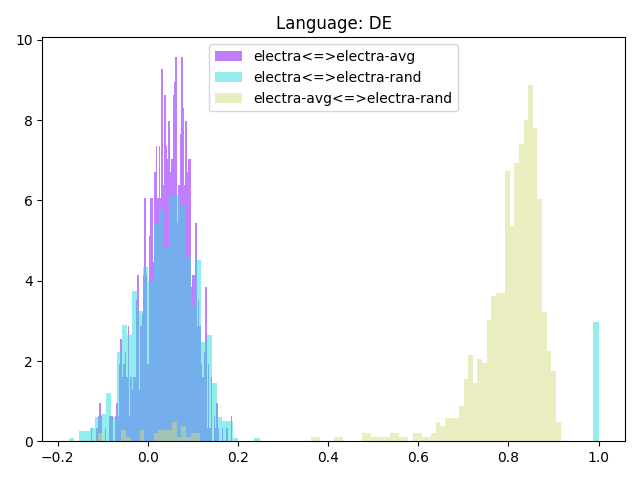} \\ \hline
    \vcent{Italian}   & \includegraphics[width=0.2\linewidth,trim={0 0 0 1cm},clip]{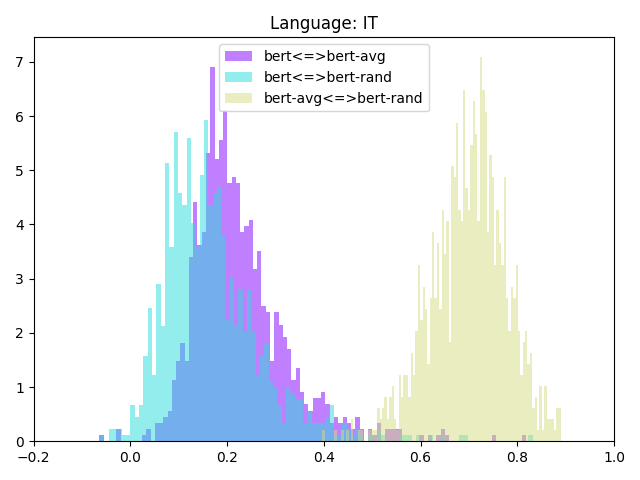} & \includegraphics[width=0.2\linewidth,trim={0 0 0 1cm},clip]{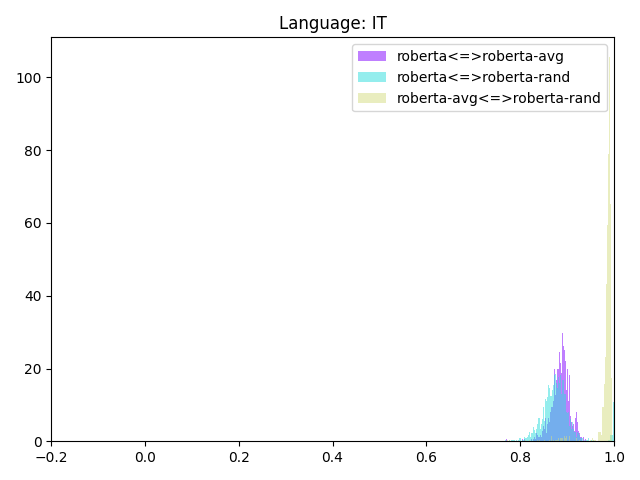} & \includegraphics[width=0.2\linewidth,trim={0 0 0 1cm},clip]{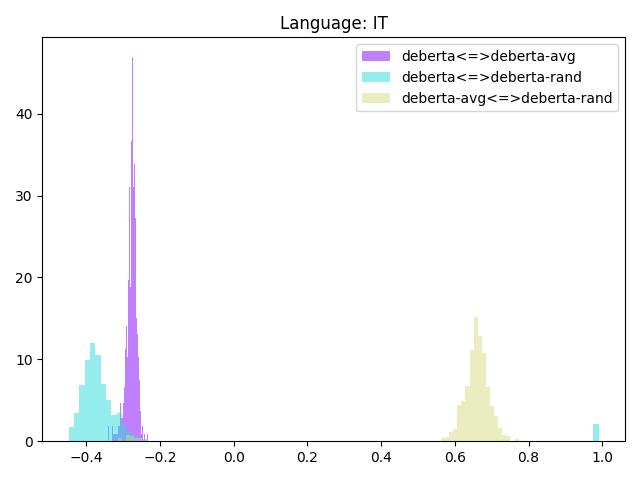} & \includegraphics[width=0.2\linewidth,trim={0 0 0 1cm},clip]{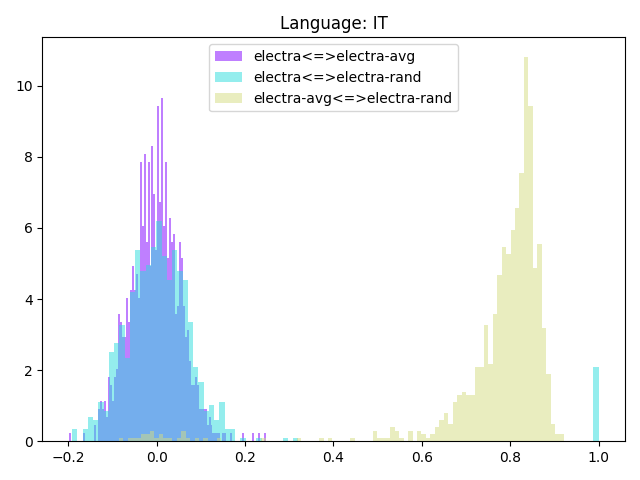} \\ \hline
    \vcent{Romanian}   & \includegraphics[width=0.2\linewidth,trim={0 0 0 1cm},clip]{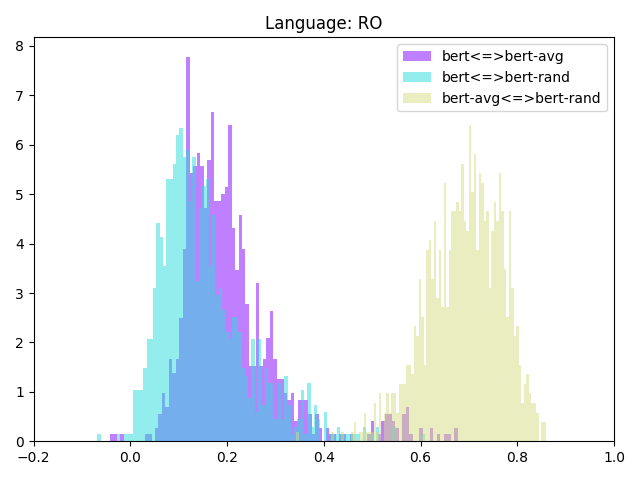} & \includegraphics[width=0.2\linewidth,trim={0 0 0 1cm},clip]{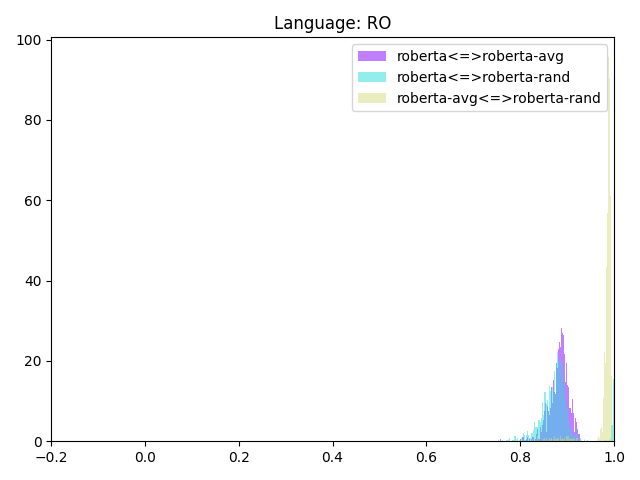} & \includegraphics[width=0.2\linewidth,trim={0 0 0 1cm},clip]{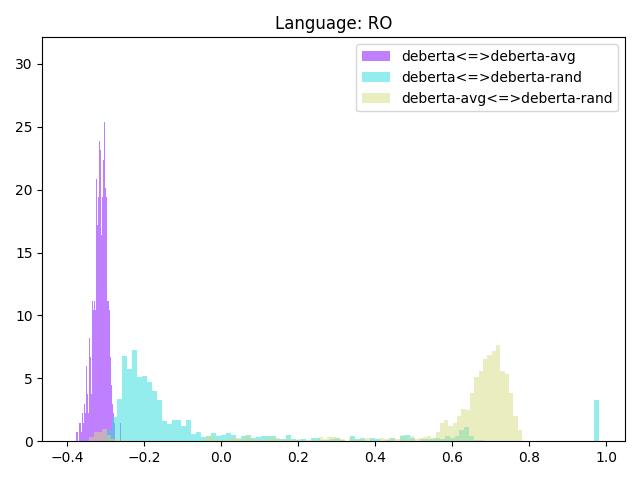} & \includegraphics[width=0.2\linewidth,trim={0 0 0 1cm},clip]{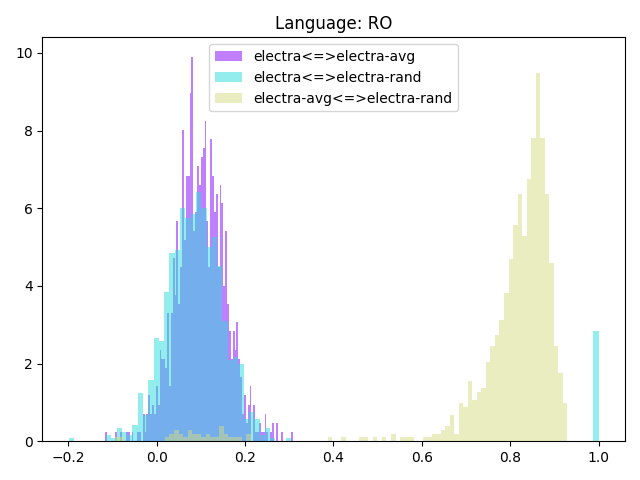} \\ \hline
    \vcent{Spanish}   & \includegraphics[width=0.2\linewidth,trim={0 0 0 1cm},clip]{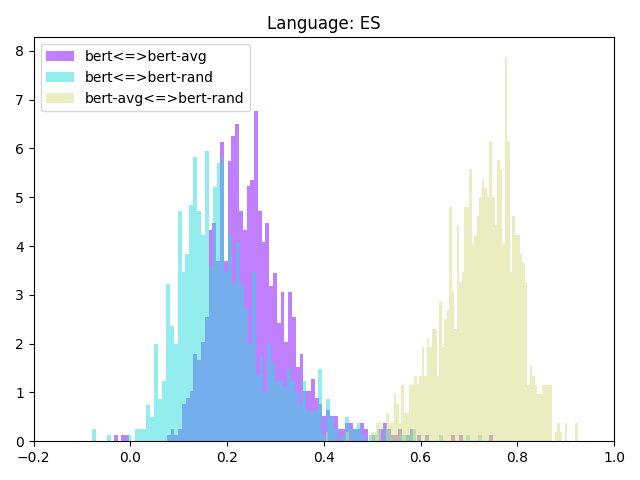} & \includegraphics[width=0.2\linewidth,trim={0 0 0 1cm},clip]{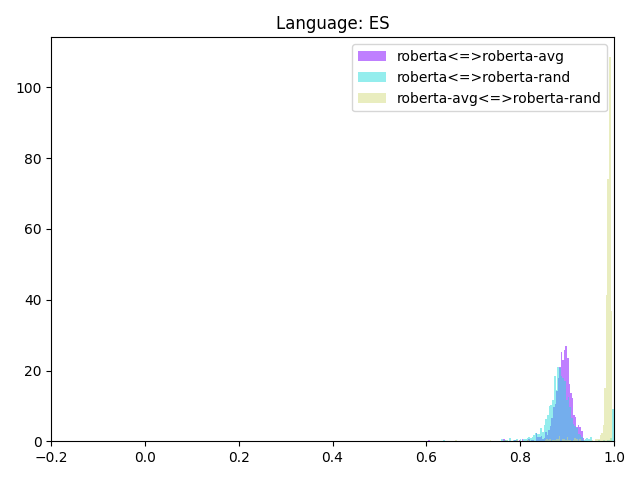} & \includegraphics[width=0.2\linewidth,trim={0 0 0 1cm},clip]{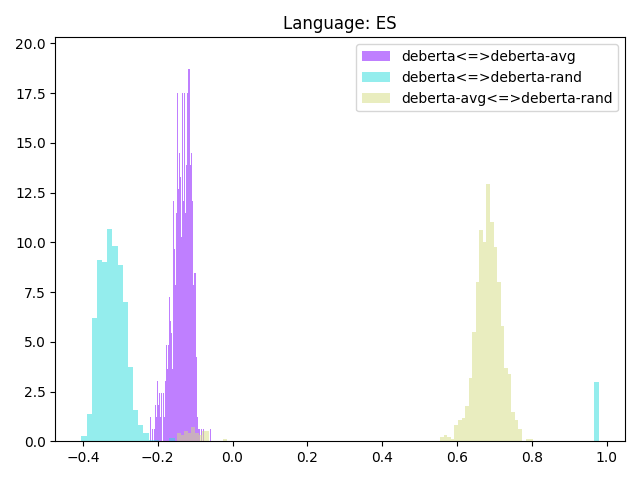} & \includegraphics[width=0.2\linewidth,trim={0 0 0 1cm},clip]{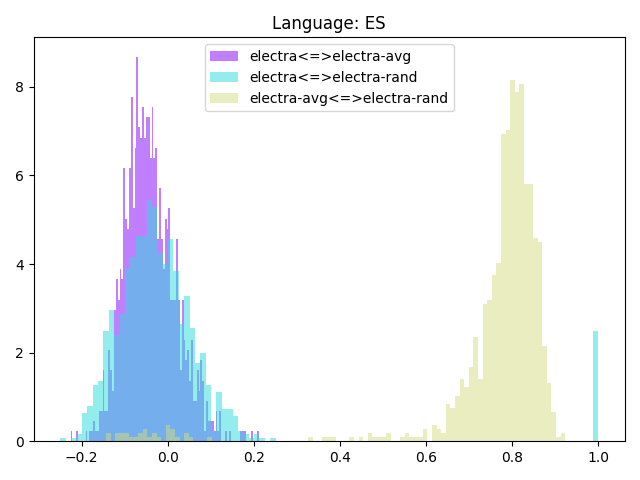} \\ \hline
    \end{tabular}
    \caption{Cosine similarities histograms computed for 1000 English/French/German/Italian/Romanian/Spanish sentences. In yellow are the distances between \savg~ and \trand, in blue the distances between \scls~ and \trand, and in purple the distances between \scls~ and \savg.}
    \label{fig:histograms_nlang}

\end{figure}

\newpage

\section{Task results}
\begin{figure}[h!]
    \centering
    {\bf Morphology tasks}
    \includegraphics[width=\linewidth,trim={0 3cm 0 1cm},clip]{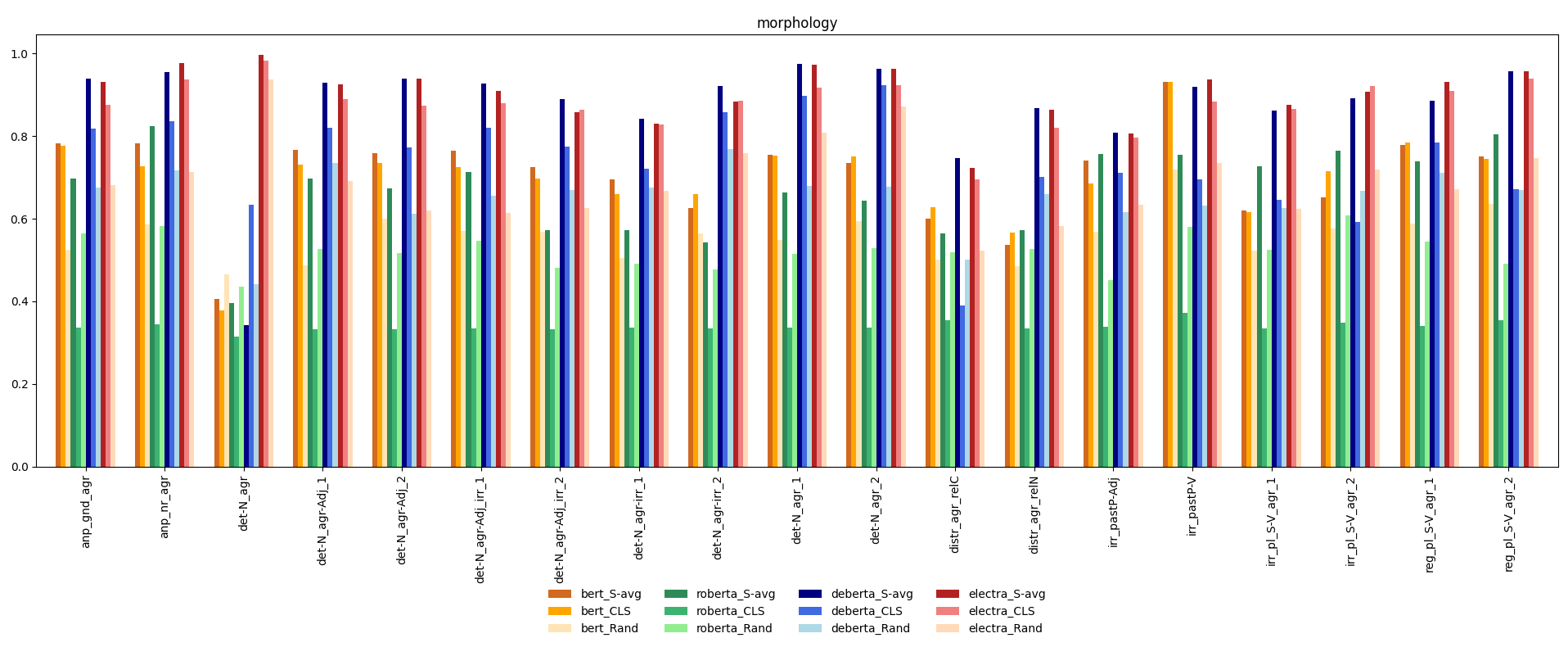}
    {\bf Syntactic tasks}
    \includegraphics[width=\linewidth,trim={0 3cm 0 1cm},clip]{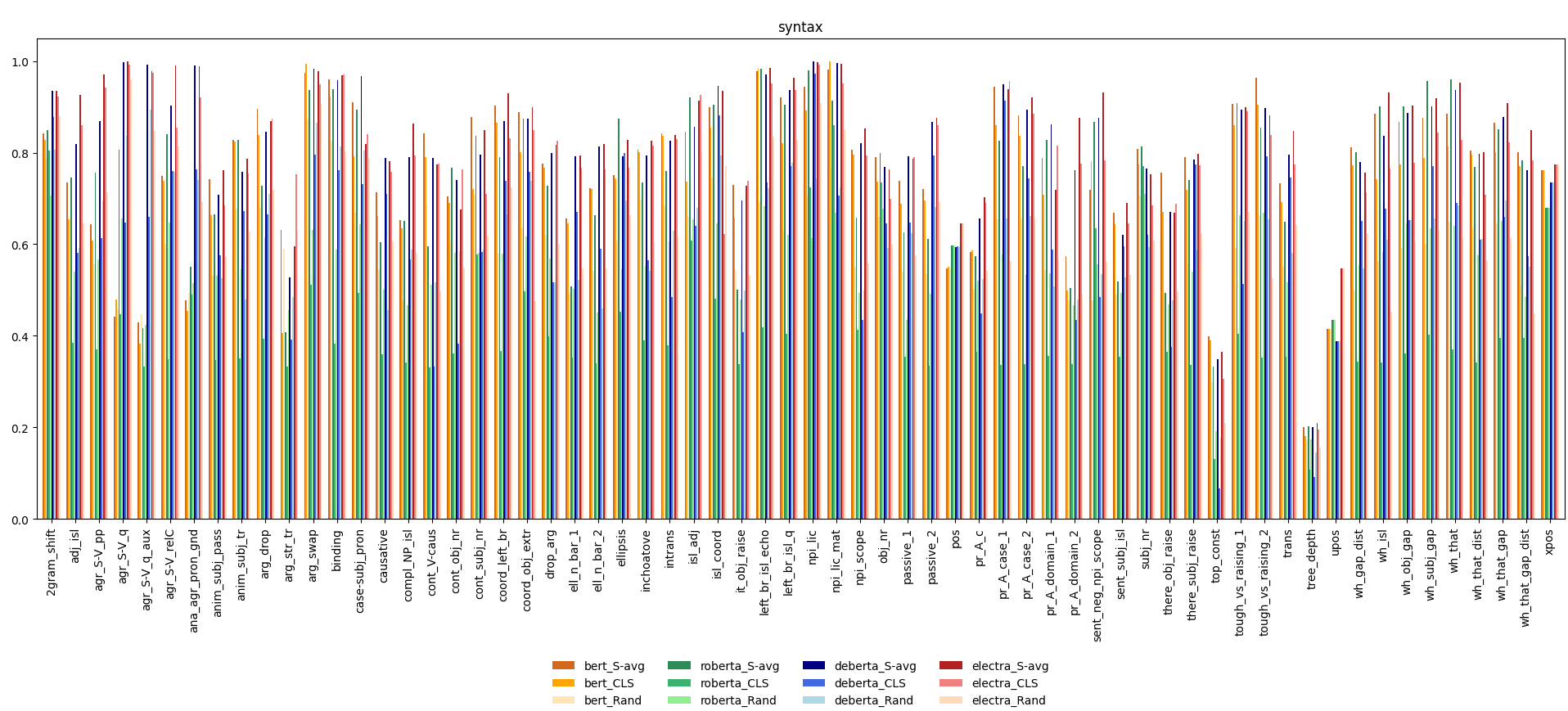}
    {\bf Semantic tasks}
    \includegraphics[width=\linewidth,trim={0 0 0 0.7cm},clip]{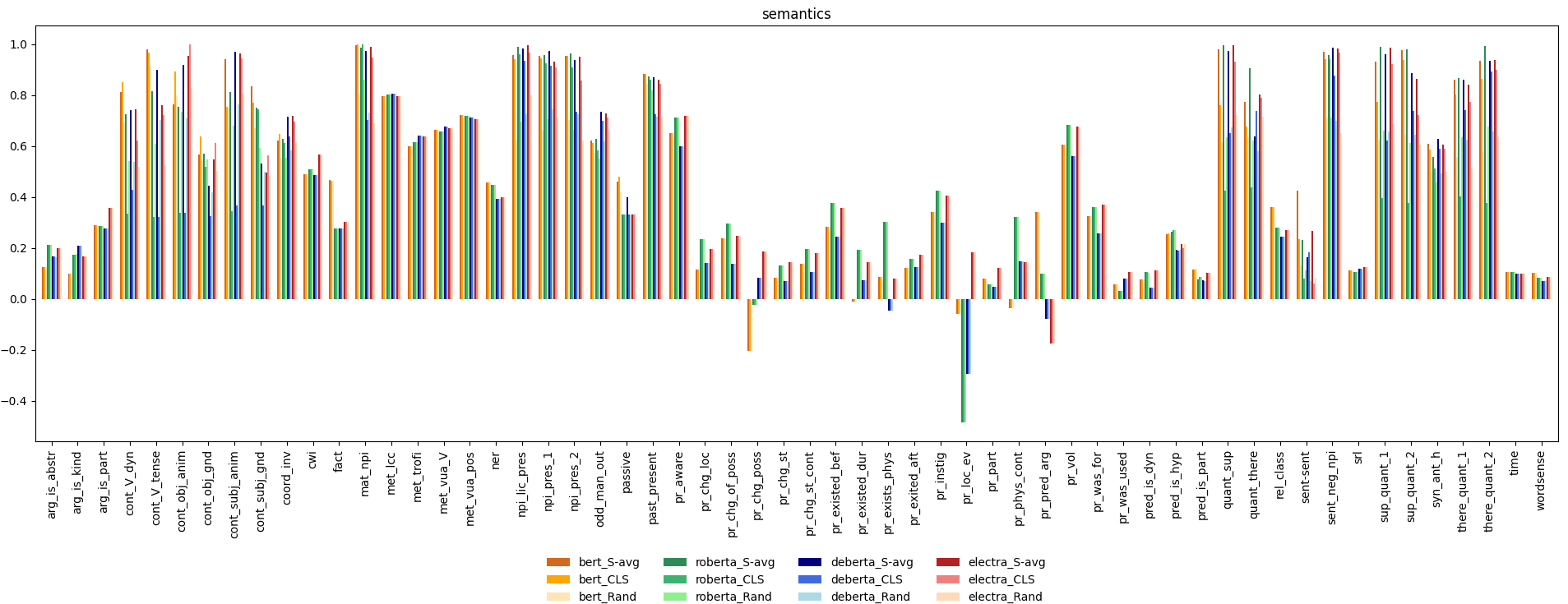}
    \caption{Detailed results on the FlashHolmes benchmark, on morphology, syntax and semantic tasks}
    \label{fig:holmes-fine-1}
\end{figure}

\begin{figure}
    \centering
    {\bf Discourse tasks}
    \includegraphics[width=\linewidth,trim={0 3cm 0 1cm},clip]{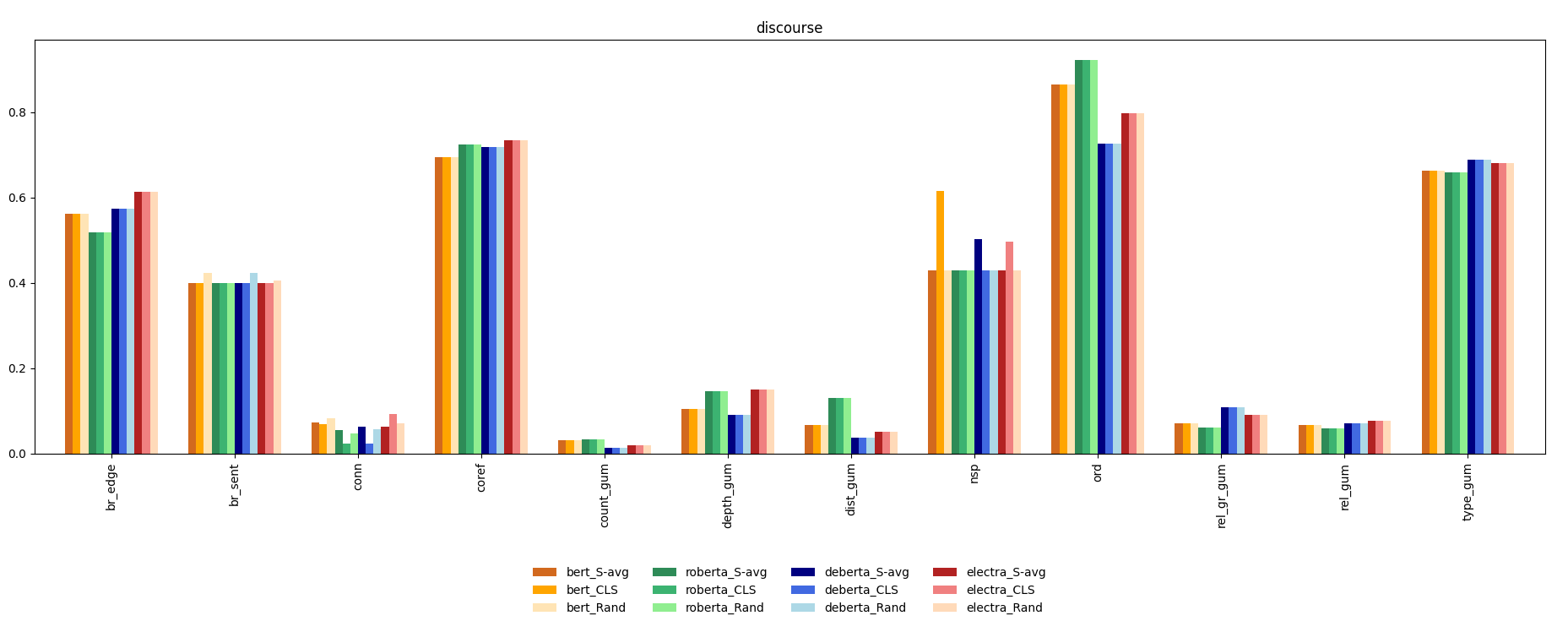}
    {\bf Reasoning tasks}
    \includegraphics[width=\linewidth,trim={0 0 0 0.7cm},clip]{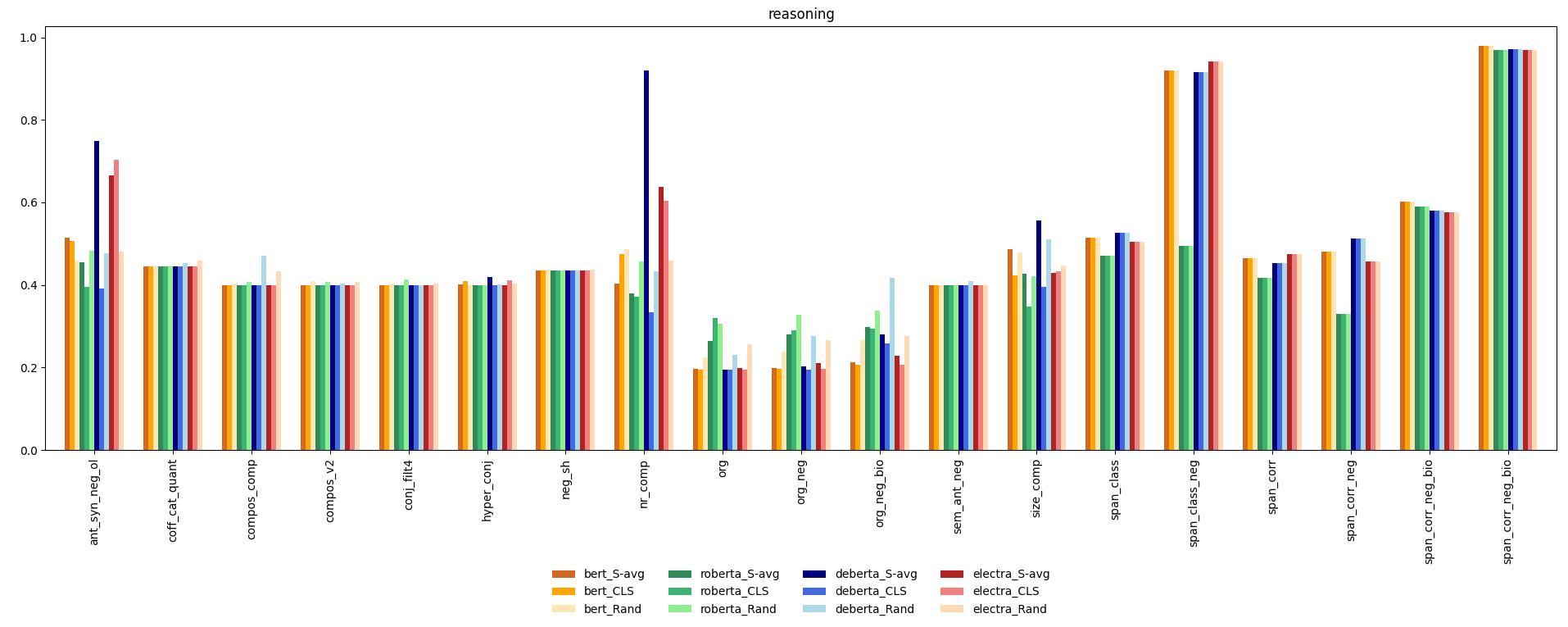}
    \caption{Detailed results on the FlashHolmes benchmark, on discourse and reasoning tasks}
    \label{fig:holmes-fine-2}
\end{figure}

\begin{landscape}
    \centering
    \begin{table}
    \begin{tabular}{l|ccc|ccc|ccc|ccc} \\
    Data & \multicolumn{3}{c}{BERT} & \multicolumn{3}{c}{RoBERTa} & \multicolumn{3}{c}{DeBERTa}  & \multicolumn{3}{c}{Electra} \\ 
  &  S-avg  &  CLS  &  Rand  &  S-avg  &  CLS  &  Rand &  S-avg  &  CLS  &  Rand &  S-avg  &  CLS  &  Rand \\ \hline
blimp-anaphor\_gender\_agreement  &   \ul{0.782}   &  0.776  &  0.525  &   \ul{0.696}   &  0.336  &  0.564  &   \textbf{0.939}   &  0.818  &  0.676  &   \ul{0.931}   &  0.875  &  0.682 \\
blimp-anaphor\_number\_agreement  &   \ul{0.782}   &  0.727  &  0.586  &   \ul{0.825}   &  0.345  &  0.582  &   \ul{0.954}   &  0.835  &  0.717  &   \textbf{0.977}   &  0.937  &  0.712 \\
blimp-determiner\_noun\_agreement\_1  &   \ul{0.755}   &  0.752  &  0.549  &   \ul{0.664}   &  0.336  &  0.514  &   \textbf{0.974}   &  0.898  &  0.678  &   \ul{0.973}   &  0.918  &  0.809 \\
blimp-determiner\_noun\_agreement\_2  &  0.736  &   \ul{0.75}  &  0.594  &  \ul{0.644}  &  0.336  &  0.529  &   \textbf{0.963}   &  0.922  &  0.677  &   \textbf{0.963}   &  0.924  &  0.873 \\
blimp-determiner\_noun\_agreement\_irregular\_1  &   \ul{0.694}   &  0.66  &  0.505  &   \ul{0.572}   &  0.336  &  0.49  &   \textbf{0.842}   &  0.721  &  0.675  &   \ul{0.83}   &  0.829  &  0.668 \\
blimp-determiner\_noun\_agreement\_irregular\_2  &  0.626  &   \ul{0.66}   &  0.565  &   \ul{0.542}   &  0.334  &  0.476  &   \textbf{0.922}   &  0.857  &  0.768  &  0.884  &   \ul{0.885}   &  0.758 \\
blimp-determiner\_noun\_agreement\_with\_adj\_2  &   \ul{0.759}   &  0.735  &  0.601  &   \ul{0.673}   &  0.333  &  0.517  &   \ul{0.938}   &  0.773  &  0.611  &   \textbf{0.939}   &  0.873  &  0.620 \\
blimp-determiner\_noun\_agreement\_with\_adj\_irregular\_1  &   \ul{0.765}   &  0.725  &  0.57  &   \ul{0.713}   &  0.334  &  0.546  &   \textbf{0.927}   &  0.82  &  0.656  &   \ul{0.909}   &  0.879  &  0.613 \\
blimp-determiner\_noun\_agreement\_with\_adj\_irregular\_2  &   \ul{0.725}   &  0.698  &  0.568  &   \ul{0.573}   &  0.333  &  0.481  &   \textbf{0.889}   &  0.775  &  0.670  &  0.857  &   \ul{0.864}   &  0.627 \\
blimp-determiner\_noun\_agreement\_with\_adjective\_1  &   \ul{0.767}   &  0.73  &  0.487  &   \ul{0.697}   &  0.333  &  0.528  &   \textbf{0.929}   &  0.82  &  0.736  &   \ul{0.925}   &  0.89  &  0.692 \\
blimp-distractor\_agreement\_relational\_noun  &  0.537  &   \ul{0.567}   &  0.485  &   \ul{0.573}   &  0.334  &  0.526  &   \textbf{0.868}   &  0.702  &  0.659  &   \ul{0.864}   &  0.819  &  0.582 \\
blimp-distractor\_agreement\_relative\_clause  &  0.6  &   \ul{0.628}   &  0.501  &   \ul{0.565}   &  0.354  &  0.518  &   \textbf{0.746}   &  0.39  &  0.501  &   \ul{0.723}   &  0.696  &  0.524 \\
blimp-irregular\_past\_participle\_adjectives  &   \ul{0.742}   &  0.685  &  0.569  &   \ul{0.757}   &  0.338  &  0.451  &   \textbf{0.808}   &  0.712  &  0.616  &   \ul{0.806}   &  0.796  &  0.633 \\
blimp-irregular\_past\_participle\_verbs  &   \ul{0.932}   &  0.931  &  0.718  &   \ul{0.754}   &  0.372  &  0.58  &   \ul{0.92}   &  0.695  &  0.633  &   \textbf{0.936}   &  0.884  &  0.736 \\
blimp-irregular\_plural\_subject\_verb\_agreement\_1  &   \ul{0.619}   &  0.615  &  0.523  &   \ul{0.727}   &  0.334  &  0.525  &   \ul{0.861}   &  0.647  &  0.626  &   \textbf{0.875}   &  0.866  &  0.624 \\
blimp-irregular\_plural\_subject\_verb\_agreement\_2  &  0.651  &   \ul{0.716}   &  0.577  &   \ul{0.764}   &  0.349  &  0.608  &   \ul{0.892}   &  0.591  &  0.667  &  0.908  &   \textbf{0.922}   &  0.720 \\
blimp-regular\_plural\_subject\_verb\_agreement\_1  &  0.779  &   \ul{0.785}   &  0.588  &   \ul{0.739}   &  0.341  &  0.544  &   \ul{0.885}   &  0.785  &  0.711  &   \textbf{0.93}   &  0.909  &  0.672 \\
blimp-regular\_plural\_subject\_verb\_agreement\_2  &   \ul{0.75}   &  0.746  &  0.637  &   \ul{0.804}   &  0.354  &  0.49  &   \textbf{0.957}   &  0.671  &  0.670  &   \textbf{0.957}   &  0.938  &  0.747 \\
zorro-agreement\_determiner\_noun-between\_neighbors  &  0.407  &  0.378  &   \ul{0.466}   &  0.395  &  0.315  &   \ul{0.435}   &  0.343  &   \ul{0.633}   &  0.442  &   \textbf{0.996}   &  0.983  &  0.937 \\ \hline   
average & 0.706  &  0.698  &  0.559  &  0.667  &  0.339  &  0.521  &  0.871  &  0.740  &  0.652  &   \textbf{0.904}   &  0.878  &  0.696 \\ \hline
    \end{tabular}
    \caption{FlashHolmes morphology tasks results. In \textbf{bold} are the overall best results, and \ul{underlined} are the results for the best performing variation for each transformer.}
    \label{tab:morphology}
    \end{table}

\newpage

\begin{longtable}{l|ccc|ccc|ccc|ccc} \\
    Data & \multicolumn{3}{c}{BERT} & \multicolumn{3}{c}{RoBERTa} & \multicolumn{3}{c}{DeBERTa}  & \multicolumn{3}{c}{Electra} \\ 
  &  S-avg  &  CLS  &  Rand  &  S-avg  &  CLS  &  Rand &  S-avg  &  CLS  &  Rand &  S-avg  &  CLS  &  Rand \\ \hline \endhead
  blimp-adjunct\_island  &   \ul{0.735}   &  0.655  &  0.487  &   \ul{0.746}   &  0.384  &  0.539  &   \ul{0.818}   &  0.581  &  0.592  &   \textbf{0.925}   &  0.861  &  0.646 \\
blimp-animate\_subject\_passive  &   \ul{0.743}   &  0.663  &  0.531  &   \ul{0.665}   &  0.346  &  0.531  &   \ul{0.709}   &  0.577  &  0.527  &   \textbf{0.761}   &  0.685  &  0.573 \\
blimp-animate\_subject\_trans  &   \ul{0.827}   &  0.824  &  0.678  &   \textbf{0.828}   &  0.351  &  0.546  &   \ul{0.759}   &  0.673  &  0.479  &   \ul{0.787}   &  0.755  &  0.628 \\
blimp-causative  &   \ul{0.714}   &  0.661  &  0.544  &   \ul{0.605}   &  0.36  &  0.501  &   \textbf{0.788}   &  0.71  &  0.456  &   \ul{0.782}   &  0.758  &  0.605 \\
blimp-complex\_NP\_island  &   \ul{0.652}   &  0.634  &  0.478  &   \ul{0.65}   &  0.341  &  0.467  &   \ul{0.791}   &  0.566  &  0.589  &   \textbf{0.863}   &  0.794  &  0.581 \\
blimp-coord\_str\_constr\_complex\_left\_branch  &   \ul{0.902}   &  0.865  &  0.58  &   \ul{0.79}   &  0.367  &  0.58  &   \ul{0.87}   &  0.739  &  0.665  &   \textbf{0.929}   &  0.831  &  0.724 \\
blimp-coord\_str\_constr\_object\_extraction  &   \ul{0.888}   &  0.801  &  0.637  &   \ul{0.874}   &  0.498  &  0.617  &   \ul{0.875}   &  0.759  &  0.739  &   \textbf{0.899}   &  0.85  &  0.476 \\
blimp-drop\_argument  &   \ul{0.775}   &  0.766  &  0.621  &   \ul{0.729}   &  0.4  &  0.569  &   \ul{0.8}   &  0.516  &  0.516  &  0.817  &   \textbf{0.827}   &  0.599 \\
blimp-ellipsis\_n\_bar\_1  &   \ul{0.656}   &  0.645  &  0.492  &   \ul{0.508}   &  0.352  &  0.503  &   \ul{0.792}   &  0.67  &  0.522  &   \textbf{0.794}   &  0.767  &  0.548 \\
blimp-ellipsis\_n\_bar\_2  &   \ul{0.723}   &  0.72  &  0.543  &   \ul{0.664}   &  0.34  &  0.45  &   \ul{0.813}   &  0.589  &  0.46  &   \textbf{0.818}   &  0.763  &  0.547 \\
blimp-existential\_there\_object\_raising  &   \textbf{0.757}   &  0.671  &  0.5  &   \ul{0.494}   &  0.364  &  0.469  &   \ul{0.67}   &  0.375  &  0.478  &  0.669  &   \ul{0.688}   &  0.497 \\
blimp-existential\_there\_subject\_raising  &   \ul{0.791}   &  0.718  &  0.558  &   \ul{0.74}   &  0.336  &  0.54  &   \ul{0.785}   &  0.774  &  0.587  &   \textbf{0.797}   &  0.772  &  0.624 \\
blimp-expletive\_it\_object\_raising  &   \ul{0.729}   &  0.659  &  0.544  &   \ul{0.5}   &  0.337  &  0.48  &   \ul{0.695}   &  0.409  &  0.499  &  0.729  &   \textbf{0.739}   &  0.532 \\
blimp-inchoative  &   \ul{0.806}   &  0.801  &  0.698  &   \ul{0.736}   &  0.39  &  0.552  &   \ul{0.793}   &  0.565  &  0.542  &   \textbf{0.825}   &  0.815  &  0.604 \\
blimp-intransitive  &   \textbf{0.841}   &  0.837  &  0.685  &   \ul{0.76}   &  0.379  &  0.606  &   \ul{0.826}   &  0.485  &  0.629  &   \ul{0.838}   &  0.829  &  0.615 \\
blimp-left\_branch\_island\_echo\_question  &  0.978  &   \ul{0.984}   &  0.692  &   \ul{0.983}   &  0.418  &  0.683  &   \ul{0.97}   &  0.735  &  0.722  &   \textbf{0.985}   &  0.95  &  0.835 \\
blimp-left\_branch\_island\_simple\_question  &   \ul{0.921}   &  0.821  &  0.63  &   \ul{0.904}   &  0.405  &  0.62  &   \ul{0.937}   &  0.771  &  0.777  &   \textbf{0.964}   &  0.937  &  0.828 \\
blimp-only\_npi\_scope  &   \ul{0.807}   &  0.796  &  0.55  &   \ul{0.658}   &  0.413  &  0.494  &   \ul{0.822}   &  0.435  &  0.498  &   \textbf{0.853}   &  0.795  &  0.559 \\
blimp-passive\_1  &   \ul{0.738}   &  0.689  &  0.541  &   \ul{0.625}   &  0.355  &  0.434  &   \textbf{0.791}   &  0.648  &  0.624  &  0.787  &   \textbf{0.791}   &  0.576 \\
blimp-passive\_2  &   \ul{0.72}   &  0.696  &  0.534  &   \ul{0.611}   &  0.335  &  0.492  &   \ul{0.867}   &  0.793  &  0.681  &   \textbf{0.876}   &  0.859  &  0.691 \\
blimp-principle\_A\_c\_command  &  0.583  &   \ul{0.586}   &  0.5  &   \ul{0.574}   &  0.365  &  0.52  &   \ul{0.657}   &  0.448  &  0.525  &   \textbf{0.702}   &  0.69  &  0.541 \\
blimp-principle\_A\_case\_1  &   \ul{0.944}   &  0.86  &  0.654  &   \ul{0.826}   &  0.336  &  0.577  &   \ul{0.949}   &  0.913  &  0.657  &  0.939  &   \textbf{0.956}   &  0.563 \\
blimp-principle\_A\_case\_2  &   \ul{0.882}   &  0.836  &  0.657  &   \ul{0.771}   &  0.338  &  0.532  &   \ul{0.894}   &  0.744  &  0.662  &   \textbf{0.921}   &  0.886  &  0.657 \\
blimp-principle\_A\_domain\_1  &   \ul{0.788}   &  0.708  &  0.544  &   \ul{0.828}   &  0.356  &  0.536  &   \textbf{0.862}   &  0.588  &  0.507  &  0.719  &   \ul{0.816}   &  0.567 \\
blimp-principle\_A\_domain\_2  &   \ul{0.575}   &  0.499  &  0.478  &   \ul{0.505}   &  0.339  &  0.468  &   \ul{0.762}   &  0.434  &  0.479  &   \textbf{0.877}   &  0.775  &  0.49 \\
blimp-principle\_A\_domain\_3  &  0.64  &   \ul{0.649}   &  0.495  &   \ul{0.524}   &  0.335  &  0.493  &   \ul{0.839}   &  0.533  &  0.515  &   \textbf{0.877}   &  0.82  &  0.548 \\
blimp-principle\_A\_reconstruction  &   \ul{0.948}   &  0.88  &  0.58  &   \ul{0.752}   &  0.333  &  0.557  &   \ul{0.808}   &  0.467  &  0.584  &   \textbf{0.949}   &  0.599  &  0.585 \\
blimp-sentential\_negation\_npi\_scope  &  0.719  &   \ul{0.781}   &  0.477  &   \ul{0.867}   &  0.634  &  0.556  &   \ul{0.876}   &  0.484  &  0.534  &   \textbf{0.931}   &  0.783  &  0.561 \\
blimp-sentential\_subject\_island  &   \ul{0.669}   &  0.645  &  0.49  &   \ul{0.519}   &  0.354  &  0.493  &   \ul{0.62}   &  0.595  &  0.527  &   \textbf{0.69}   &  0.645  &  0.534 \\
blimp-tough\_vs\_raising\_1  &   \ul{0.907}   &  0.86  &  0.592  &   \textbf{0.908}   &  0.404  &  0.663  &   \ul{0.893}   &  0.513  &  0.651  &   \ul{0.899}   &  0.891  &  0.671 \\
blimp-tough\_vs\_raising\_2  &   \textbf{0.964}   &  0.904  &  0.661  &   \ul{0.855}   &  0.353  &  0.669  &   \ul{0.898}   &  0.792  &  0.655  &   \ul{0.882}   &  0.839  &  0.526 \\
blimp-transitive  &   \ul{0.732}   &  0.693  &  0.553  &   \ul{0.65}   &  0.354  &  0.516  &   \ul{0.796}   &  0.746  &  0.581  &   \textbf{0.847}   &  0.774  &  0.641 \\
blimp-wh\_island  &   \ul{0.886}   &  0.742  &  0.563  &   \ul{0.901}   &  0.342  &  0.583  &   \ul{0.836}   &  0.679  &  0.609  &   \textbf{0.931}   &  0.766  &  0.452 \\
blimp-wh\_questions\_object\_gap  &   \ul{0.867}   &  0.774  &  0.593  &   \ul{0.901}   &  0.361  &  0.645  &   \ul{0.888}   &  0.653  &  0.653  &   \textbf{0.902}   &  0.779  &  0.607 \\
blimp-wh\_questions\_subj\_gap  &   \ul{0.876}   &  0.789  &  0.6  &   \textbf{0.957}   &  0.402  &  0.635  &   \ul{0.9}   &  0.771  &  0.657  &   \ul{0.919}   &  0.844  &  0.63 \\
blimp-wh\_questions\_subj\_gap\_long\_dist  &   \textbf{0.812}   &  0.772  &  0.498  &   \ul{0.802}   &  0.344  &  0.555  &   \ul{0.78}   &  0.651  &  0.548  &   \ul{0.757}   &  0.714  &  0.625 \\
blimp-wh\_vs\_that\_no\_gap  &   \ul{0.885}   &  0.813  &  0.649  &   \textbf{0.959}   &  0.371  &  0.641  &   \ul{0.936}   &  0.689  &  0.685  &   \ul{0.952}   &  0.827  &  0.615 \\
blimp-wh\_vs\_that\_no\_gap\_long\_dist  &   \textbf{0.805}   &  0.796  &  0.634  &   \ul{0.769}   &  0.341  &  0.576  &   \ul{0.797}   &  0.609  &  0.612  &   \ul{0.801}   &  0.708  &  0.565 \\
blimp-wh\_vs\_that\_with\_gap  &   \ul{0.865}   &  0.802  &  0.645  &   \ul{0.851}   &  0.395  &  0.651  &   \ul{0.878}   &  0.66  &  0.696  &   \textbf{0.908}   &  0.822  &  0.613 \\
blimp-wh\_vs\_that\_with\_gap\_long\_dist  &   \ul{0.8}   &  0.771  &  0.511  &   \ul{0.784}   &  0.394  &  0.486  &   \ul{0.762}   &  0.574  &  0.55  &   \textbf{0.849}   &  0.783  &  0.449 \\
const  &  0.163  &  0.163  &   \ul{0.164} &  \ul{0.128} &  \ul{0.128}  &  \ul{0.128}  &   \textbf{0.186}   &   \textbf{0.186}   &   \textbf{0.186}   &  \ul{0.18}  &  \ul{0.18}  &  \ul{0.18}  \\
const\_max\_depth  &    &   \ul{0.815}   &  0.668  &    &   \ul{0.78}   &  0.658  &    &   \ul{0.812}   &  0.539  &    &   \textbf{0.829}   &  0.635 \\
const\_node\_length  &    &   \ul{0.879}   &  0.721  &    &   \ul{0.89}   &  0.73  &    &   \textbf{0.903}   &  0.588  &    &   \ul{0.871}   &  0.676 \\
context-object\_number  &   \ul{0.705}   &  0.691  &  0.609  &   \textbf{0.767}   &  0.361  &  0.581  &   \ul{0.74}   &  0.383  &  0.586  &  0.676  &   \ul{0.763}   &  0.548 \\
context-subj\_number  &   \textbf{0.878}   &  0.72  &  0.707  &   \ul{0.837}   &  0.578  &  0.582  &   \ul{0.795}   &  0.582  &  0.588  &   \ul{0.849}   &  0.71  &  0.619 \\
context-verb\_causative  &   \textbf{0.842}   &  0.791  &  0.687  &   \ul{0.596}   &  0.332  &  0.511  &   \ul{0.789}   &  0.333  &  0.517  &  0.775  &   \ul{0.777}   &  0.498 \\
flesch  &    &   \ul{0.171}   &  0.019  &    &   \ul{0.605}   &  0.082  &    &   \textbf{0.846}   &  0.002  &    &   \ul{0.48}   &  -0.074 \\
pos  &  0.547  &   \ul{0.551}   &  0.548  &   \ul{0.598}   &  0.597  &   \ul{0.598}   &  0.594  &   \ul{0.595}   &  0.594  &   \textbf{0.646}   &  0.645  &  0.645 \\
senteval-bigram\_shift  &   \ul{0.841}   &  0.828  &  0.786  &   \ul{0.85}   &  0.804  &  0.706  &   \textbf{0.936}   &  0.878  &  0.806  &   \ul{0.935}   &  0.922  &  0.88 \\
senteval-obj\_number  &   \ul{0.791}   &  0.737  &  0.66  &   \textbf{0.799}   &  0.735  &  0.678  &   \ul{0.769}   &  0.645  &  0.592  &   \ul{0.763}   &  0.698  &  0.599 \\
senteval-sentence\_length  &   \textbf{0.048}   &   \textbf{0.048}   &   \textbf{0.048}   &   \textbf{0.048}   &   \textbf{0.048}   &   \textbf{0.048}   &   \textbf{0.048}   &   \textbf{0.048}   &   \textbf{0.048}   &   \textbf{0.048}   &   \textbf{0.048}   &   \textbf{0.048} \\
senteval-subj\_number  &   \ul{0.808}   &  0.774  &  0.707  &   \textbf{0.813}   &  0.771  &  0.71  &   \ul{0.766}   &  0.62  &  0.593  &   \ul{0.752}   &  0.685  &  0.608 \\
senteval-top\_constituents  &   \textbf{0.398}   &  0.391  &  0.298  &   \ul{0.334}   &  0.131  &  0.191  &   \ul{0.349}   &  0.067  &  0.177  &   \ul{0.365}   &  0.307  &  0.209 \\
senteval-tree\_depth  &   \ul{0.2}   &  0.181  &  0.174  &   \ul{0.203}   &  0.108  &  0.173  &   \ul{0.2}   &  0.091  &  0.145  &   \textbf{0.209}   &  0.196  &  0.156 \\
upos  &   \ul{0.415}   &  \ul{0.415}  & \ul{0.415}  &  \ul{0.434} &  \ul{0.434}  &  \ul{0.434}  &  \ul{0.388}  &  \ul{0.388}  &  \ul{0.388}   &   \textbf{0.548}   &   \textbf{0.548}   &   \textbf{0.548} \\
xpos  &   \ul{0.761}  &  \ul{0.761}  &  \ul{0.761}  &  \ul{0.68} &  \ul{0.68} & \ul{0.68}  &  \ul{0.736} &  0.735  &  \ul{0.736}  &   \textbf{0.775}   &   \textbf{0.775}   &   \textbf{0.775}  \\
zorro-agr\_subj\_v-across\_prepositional\_phrase  &   \ul{0.643}   &  0.608  &  0.556  &   \ul{0.757}   &  0.371  &  0.567  &   \ul{0.868}   &  0.614  &  0.694  &   \textbf{0.971}   &  0.943  &  0.714 \\
zorro-agr\_subj\_v-across\_relative\_clause  &   \ul{0.75}   &  0.738  &  0.6  &   \ul{0.84}   &  0.348  &  0.647  &   \ul{0.904}   &  0.76  &  0.758  &   \textbf{0.99}   &  0.855  &  0.813 \\
zorro-agr\_subj\_v-in\_question\_with\_aux  &  0.429  &  0.383  &   \ul{0.447}   &  0.417  &  0.333  &   \ul{0.423}   &   \textbf{0.992}   &  0.659  &  0.894  &   \ul{0.978}   &  0.975  &  0.847 \\
zorro-agr\_subj\_v-in\_simple\_question  &  0.441  &   \ul{0.48}   &  0.457  &   \ul{0.806}   &  0.447  &  0.656  &   \ul{0.998}   &  0.648  &  0.836  &   \textbf{1.0}   &  0.991  &  0.96 \\
zorro-anaphor\_agr-pronoun\_gender  &  0.477  &  0.454  &   \ul{0.494}   &   \ul{0.551}   &  0.49  &  0.516  &   \textbf{0.991}   &  0.764  &  0.74  &   \ul{0.989}   &  0.921  &  0.691 \\
zorro-arg\_str-dropped\_argument  &   \textbf{0.896}   &  0.838  &  0.68  &   \ul{0.728}   &  0.393  &  0.615  &   \ul{0.845}   &  0.665  &  0.71  &  0.869  &   \ul{0.875}   &  0.719 \\
zorro-arg\_str-swapped\_arguments  &  0.974  &   \textbf{0.994}   &  0.874  &   \ul{0.936}   &  0.512  &  0.63  &   \ul{0.983}   &  0.795  &  0.865  &   \ul{0.978}   &  0.95  &  0.906 \\
zorro-arg\_str-transitive  &   \ul{0.63}   &  0.405  &  0.59  &  0.409  &  0.333  &   \ul{0.455}   &   \ul{0.527}   &  0.392  &  0.484  &  0.596  &   \textbf{0.754}   &  0.63 \\
zorro-binding-principle\_a  &   \ul{0.96}   &  0.923  &  0.827  &   \ul{0.939}   &  0.383  &  0.588  &   \ul{0.958}   &  0.762  &  0.813  &  0.968  &   \textbf{0.97}   &  0.804 \\
zorro-case-subjective\_pronoun  &   \ul{0.91}   &  0.793  &  0.669  &   \ul{0.895}   &  0.493  &  0.643  &   \textbf{0.967}   &  0.731  &  0.805  &  0.819  &   \ul{0.841}   &  0.787 \\
zorro-ellipsis-n\_bar  &   \ul{0.751}   &  0.744  &  0.608  &   \textbf{0.874}   &  0.453  &  0.545  &  0.792  &   \ul{0.799}   &  0.696  &   \ul{0.829}   &  0.767  &  0.663 \\
zorro-filler-gap-wh\_question\_object  &   \ul{0.959}   &  0.848  &  0.769  &   \textbf{0.997}   &  0.785  &  0.774  &   \ul{0.909}   &  0.781  &  0.743  &   \ul{0.967}   &  0.915  &  0.756 \\
zorro-filler-gap-wh\_question\_subject  &   \ul{0.99}   &  0.904  &  0.827  &   \textbf{1.0}   &  0.902  &  0.673  &   \ul{0.942}   &  0.797  &  0.812  &  0.912  &   \ul{0.933}   &  0.786 \\
zorro-island-effects-adjunct\_island  &   \ul{0.846}   &  0.738  &  0.659  &   \ul{0.921}   &  0.607  &  0.654  &   \ul{0.857}   &  0.641  &  0.679  &  0.914  &   \textbf{0.926}   &  0.799 \\
zorro-island-effects-coord\_str\_constr  &   \ul{0.899}   &  0.855  &  0.746  &   \ul{0.904}   &  0.482  &  0.646  &   \textbf{0.946}   &  0.882  &  0.795  &   \ul{0.936}   &  0.623  &  0.771 \\
zorro-local\_attractor-in\_question\_with\_aux  &   \ul{0.669}   &  0.585  &  0.525  &   \ul{0.807}   &  0.399  &  0.642  &   \ul{0.878}   &  0.645  &  0.729  &   \textbf{0.969}   &  0.908  &  0.807 \\
zorro-npi\_licensing-matrix\_question  &  0.981  &   \textbf{0.999}   &  0.767  &   \ul{0.914}   &  0.859  &  0.668  &   \ul{0.996}   &  0.706  &  0.723  &   \ul{0.994}   &  0.951  &  0.851 \\
zorro-npi\_licensing-only\_npi\_licensor  &   \ul{0.944}   &  0.892  &  0.704  &   \ul{0.98}   &  0.725  &  0.614  &   \textbf{1.0}   &  0.972  &  0.744  &   \ul{0.998}   &  0.992  &  0.909 \\ \hline
average & 0.752  &  0.705  &  0.577  &  0.723  &  0.435  &  0.547  &  0.795  &  0.622  &  0.596  &   \textbf{0.819}   &  0.774  &  0.615 \\ \hline
    \caption{FlashHolmes syntactic tasks results. In \textbf{bold} are the overall best results, and \ul{underlined} are the results for the best performing variation for each transformer.}
    \label{tab:syntax}
\end{longtable}

\newpage

\begin{longtable}{l|ccc|ccc|ccc|ccc} \\
    Data & \multicolumn{3}{c}{BERT} & \multicolumn{3}{c}{RoBERTa} & \multicolumn{3}{c}{DeBERTa}  & \multicolumn{3}{c}{Electra} \\ 
  &  S-avg  &  CLS  &  Rand  &  S-avg  &  CLS  &  Rand &  S-avg  &  CLS  &  Rand &  S-avg  &  CLS  &  Rand \\ \hline \endhead
arg-is-abstract  &   \ul{0.125}   &   \ul{0.125}   &   \ul{0.125}   &   \textbf{0.212}   &   \textbf{0.212}   &   \textbf{0.212}   &   \ul{0.167}   &   \ul{0.167}   &  0.165  &   \ul{0.2}   &   \ul{0.2}   &   \ul{0.2} \\
arg-is-kind  &   \ul{0.098}   &   \ul{0.098}   &   \ul{0.098}   &  0.174  &  0.175  &   \ul{0.176}   &   \textbf{0.209}   &  0.208  &   \textbf{0.209}   &   \ul{0.168}   &   \ul{0.168}   &   \ul{0.168} \\
arg-is-particular  &   \ul{0.29}   &   \ul{0.29}   &   \ul{0.29}   &   \ul{0.286}   &  0.285  &  0.285  &  0.277  &  0.277  &   \ul{0.278}   &   \textbf{0.359}   &   \textbf{0.359}   &   \textbf{0.359} \\
blimp-existential\_there\_quantifiers\_1  &   \ul{0.861}   &  0.802  &  0.556  &   \textbf{0.868}   &  0.402  &  0.634  &   \ul{0.859}   &  0.741  &  0.627  &   \ul{0.841}   &  0.775  &  0.512 \\
blimp-existential\_there\_quantifiers\_2  &   \ul{0.936}   &  0.863  &  0.578  &   \textbf{0.993}   &  0.378  &  0.677  &   \ul{0.934}   &  0.894  &  0.659  &   \ul{0.939}   &  0.898  &  0.639 \\
blimp-matrix\_quest\_npi\_licensor\_pres  &  0.995  &   \textbf{1.0}   &  0.807  &  0.986  &   \textbf{1.0}   &  0.862  &   \ul{0.973}   &  0.704  &  0.733  &   \ul{0.989}   &  0.948  &  0.691 \\
blimp-npi\_present\_1  &   \ul{0.956}   &  0.946  &  0.66  &   \ul{0.957}   &  0.927  &  0.706  &   \textbf{0.973}   &  0.915  &  0.746  &   \ul{0.93}   &  0.91  &  0.711 \\
blimp-npi\_present\_2  &  0.953  &   \ul{0.954}   &  0.702  &   \textbf{0.965}   &  0.91  &  0.664  &   \ul{0.939}   &  0.737  &  0.727  &   \ul{0.95}   &  0.858  &  0.622 \\
blimp-only\_npi\_licensor\_present  &   \ul{0.959}   &  0.941  &  0.625  &   \ul{0.99}   &  0.962  &  0.698  &   \ul{0.982}   &  0.934  &  0.726  &   \textbf{0.995}   &  0.966  &  0.8 \\
blimp-sent\_neg\_npi\_licensor\_present  &   \ul{0.972}   &  0.942  &  0.712  &   \ul{0.956}   &  0.943  &  0.711  &   \textbf{0.987}   &  0.877  &  0.699  &   \ul{0.985}   &  0.966  &  0.652 \\
blimp-superlative\_quantifiers\_1  &   \ul{0.932}   &  0.773  &  0.626  &   \textbf{0.99}   &  0.398  &  0.661  &   \ul{0.96}   &  0.621  &  0.657  &   \ul{0.986}   &  0.922  &  0.687 \\
blimp-superlative\_quantifiers\_2  &   \ul{0.976}   &  0.939  &  0.645  &   \textbf{0.98}   &  0.378  &  0.612  &   \ul{0.888}   &  0.738  &  0.643  &   \ul{0.863}   &  0.723  &  0.606 \\
context-object\_animacy  &  0.765  &   \ul{0.894}   &  0.8  &   \ul{0.753}   &  0.339  &  0.736  &   \ul{0.918}   &  0.339  &  0.71  &  0.955  &   \textbf{0.999}   &  0.83 \\
context-object\_gender  &  0.566  &   \textbf{0.638}   &  0.543  &   \ul{0.57}   &  0.518  &  0.548  &   \ul{0.444}   &  0.326  &  0.421  &  0.547  &   \ul{0.613}   &  0.504 \\
context-subject\_animacy  &   \ul{0.942}   &  0.756  &  0.752  &   \ul{0.814}   &  0.346  &  0.679  &   \textbf{0.969}   &  0.367  &  0.765  &   \ul{0.966}   &  0.945  &  0.806 \\
context-subject\_gender  &   \textbf{0.835}   &  0.77  &  0.671  &   \ul{0.75}   &  0.744  &  0.594  &   \ul{0.533}   &  0.367  &  0.501  &  0.496  &   \ul{0.566}   &  0.484 \\
context-verb\_dynamic  &  0.813  &   \textbf{0.852}   &  0.691  &   \ul{0.724}   &  0.334  &  0.542  &   \ul{0.742}   &  0.43  &  0.54  &   \ul{0.743}   &  0.623  &  0.519 \\
context-verb\_tense  &   \textbf{0.981}   &  0.969  &  0.913  &   \ul{0.816}   &  0.324  &  0.61  &   \ul{0.898}   &  0.324  &  0.702  &   \ul{0.762}   &  0.723  &  0.525 \\
cwi  &   \ul{0.49}   &   \ul{0.49}   &   \ul{0.49}   &   \ul{0.51}   &   \ul{0.51}   &   \ul{0.51}   &   \ul{0.488}   &   \ul{0.488}   &   \ul{0.488}   &   \textbf{0.568}   &   \textbf{0.568}   &   \textbf{0.568}  \\
event\_structure-distributive  &   \ul{0.678}   &   \ul{0.678}   &   \ul{0.678}   &   \ul{0.675}   &   \ul{0.675}   &   \ul{0.675}   &   \ul{0.691}   &   \ul{0.691}   &   \ul{0.691}   &   \textbf{0.694}   &   \textbf{0.694}   &   \textbf{0.694}  \\
event\_structure-event  &   \textbf{0.241}   &   \textbf{0.241}   &   \textbf{0.241}   &   \textbf{0.241}   &   \textbf{0.241}   &   \textbf{0.241}   &   \textbf{0.241}   &   \textbf{0.241}   &   \textbf{0.241}   &   \textbf{0.241}   &   \textbf{0.241}   &   \textbf{0.241} \\
event\_structure-has-natural-parts  &   \textbf{0.452}   &   \textbf{0.452}   &   \textbf{0.452}   &   \textbf{0.452}   &   \textbf{0.452}   &   \textbf{0.452}   &   \textbf{0.452}   &   \textbf{0.452}   &   \textbf{0.452}   &   \textbf{0.452}   &   \textbf{0.452}   &   \textbf{0.452} \\
event\_structure-has-similar-parts  &   \ul{0.302}   &   \ul{0.302}   &   \ul{0.302}   &   \ul{0.284}   &   \ul{0.284}   &   \ul{0.284}   &   \ul{0.337}   &   \ul{0.337}   &   \ul{0.337}   &   \textbf{0.423}   &   \textbf{0.423}   &   \textbf{0.423} \\
event\_structure-is-dynamic  &   \ul{0.364}   &   \ul{0.364}   &   \ul{0.364}   &   \ul{0.363}   &   \ul{0.363}   &   \ul{0.363}   &   \textbf{0.395}   &   \textbf{0.395}   &   \textbf{0.395}   &   \ul{0.376}   &   \ul{0.376}   &   \ul{0.376} \\
event\_structure-is-telic  &   \textbf{0.466}   &   \textbf{0.466}   &   \textbf{0.466}   &   \textbf{0.466}   &   \textbf{0.466}   &   \textbf{0.466}   &   \textbf{0.466}   &   \textbf{0.466}   &   \textbf{0.466}   &   \textbf{0.466}   &   \textbf{0.466}   &   \textbf{0.466} \\
factuality  &   \textbf{0.466}   &   \textbf{0.466}   &   \textbf{0.466}   &   \ul{0.278}   &   \ul{0.278}   &   \ul{0.278}   &   \ul{0.278}   &   \ul{0.278}   &   \ul{0.278}   &   \ul{0.303}   &   \ul{0.303}   &   \ul{0.303} \\
metaphor-lcc  &   \ul{0.796}   &   \ul{0.796}   &   \ul{0.796}   &   \ul{0.803}   &   \ul{0.803}   &   \ul{0.803}   &   \textbf{0.806}   &   \textbf{0.806}   &   \textbf{0.806}   &   \ul{0.798}   &   \ul{0.798}   &   \ul{0.798} \\
metaphor-trofi  &   \ul{0.601}   &   \ul{0.601}   &   \ul{0.601}   &   \ul{0.617}   &   \ul{0.617}   &   \ul{0.617}   &   \textbf{0.642}   &   \textbf{0.642}   &   \textbf{0.642}   &   \ul{0.638}   &   \ul{0.638}   &   \ul{0.638} \\
metaphor-vua\_pos  &   \textbf{0.721}   &   \textbf{0.721}   &   \textbf{0.721}   &   \ul{0.719}   &   \ul{0.719}   &   \ul{0.719}   &   \ul{0.712}   &   \ul{0.712}   &   \ul{0.712}   &   \ul{0.708}   &   \ul{0.708}   &   \ul{0.708} \\
metaphor-vua\_verb  &   \ul{0.665}   &   \ul{0.665}   &   \ul{0.665}   &   \ul{0.657}   &   \ul{0.657}   &   \ul{0.657}   &   \textbf{0.678}   &   \textbf{0.678}   &   \textbf{0.678}   &   \ul{0.672}   &   \ul{0.672}   &   \ul{0.672} \\
ner  &   \textbf{0.457}   &  0.456  &   \textbf{0.457}   &   \ul{0.449}   &   \ul{0.449}   &   \ul{0.449}   &   \ul{0.393}   &   \ul{0.393}   &   \ul{0.393}   &   \ul{0.401}   &   \ul{0.401}   &   \ul{0.401} \\
passive  &  0.46  &   \textbf{0.48}   &  0.418  &   \ul{0.333}   &   \ul{0.333}   &   \ul{0.333}   &   \ul{0.399}   &  0.333  &  0.333  &  0.333  &  0.333  &   \ul{0.334} \\
pred-is-dynamic  &   \ul{0.078}   &   \ul{0.078}   &  0.075  &   \ul{0.105}   &   \ul{0.105}   &   \ul{0.105}   &   \ul{0.046}   &   \ul{0.046}   &   \ul{0.046}   &   \textbf{0.111}   &   \textbf{0.111}   &   \textbf{0.111} \\
pred-is-hypothetical  &  0.256  &   \ul{0.257}   &  0.256  &  0.264  &   \textbf{0.271}   &   \textbf{0.271}   &  0.193  &  0.191  &   \ul{0.195}   &   \ul{0.216}   &  0.199  &   \ul{0.216} \\
pred-is-particular  &  0.117  &  0.116  &   \textbf{0.118}   &  0.076  &   \ul{0.086}   &   \ul{0.086}   &   \ul{0.074}   &  0.071  &  0.071  &  0.102  &   \ul{0.104}   &  0.103 \\
protoroles-awareness  &   \ul{0.651}   &   \ul{0.651}   &   \ul{0.651}   &   \ul{0.714}   &   \ul{0.714}   &   \ul{0.714}   &   \ul{0.599}   &   \ul{0.599}   &   \ul{0.599}   &   \textbf{0.72}   &   \textbf{0.72}   &   \textbf{0.72} \\
protoroles-change\_of\_location  &   \ul{0.115}   &   \ul{0.115}   &   \ul{0.115}   &   \textbf{0.235}   &   \textbf{0.235}   &   \textbf{0.235}   &   \ul{0.141}   &   \ul{0.141}   &   \ul{0.141}   &   \ul{0.197}   &   \ul{0.197}   &   \ul{0.197} \\
protoroles-change\_of\_possession  &   \ul{0.238}   &   \ul{0.238}   &   \ul{0.238}   &   \textbf{0.298}   &   \textbf{0.298}   &   \textbf{0.298}   &   \ul{0.139}   &   \ul{0.139}   &   \ul{0.139}   &   \ul{0.25}   &   \ul{0.25}   &   \ul{0.25} \\
protoroles-change\_of\_state  &   \ul{0.082}   &   \ul{0.082}   &   \ul{0.082}   &   \ul{0.132}   &   \ul{0.132}   &   \ul{0.132}   &   \ul{0.071}   &   \ul{0.071}   &   \ul{0.071}   &   \textbf{0.144}   &   \textbf{0.144}   &   \textbf{0.144} \\
protoroles-change\_of\_state\_continuous  &   \ul{0.138}   &   \ul{0.138}   &   \ul{0.138}   &   \textbf{0.197}   &   \textbf{0.197}   &   \textbf{0.197}   &   \ul{0.107}   &   \ul{0.107}   &   \ul{0.107}   &   \ul{0.179}   &   \ul{0.179}   &   \ul{0.179} \\
protoroles-changes\_possession  &   \ul{-0.203}   &   \ul{-0.203}   &   \ul{-0.203}   &   \ul{-0.024}   &   \ul{-0.024}   &   \ul{-0.024}   &   \ul{0.085}   &   \ul{0.085}   &   \ul{0.085}   &   \textbf{0.185}   &   \textbf{0.185}   &   \textbf{0.185} \\
protoroles-existed\_after  &   \ul{0.122}   &   \ul{0.122}   &   \ul{0.122}   &   \ul{0.157}   &   \ul{0.157}   &   \ul{0.157}   &   \ul{0.126}   &   \ul{0.126}   &   \ul{0.126}   &   \textbf{0.173}   &   \textbf{0.173}   &   \textbf{0.173} \\
protoroles-existed\_before  &   \ul{0.283}   &   \ul{0.283}   &   \ul{0.283}   &   \textbf{0.377}   &   \textbf{0.377}   &   \textbf{0.377}   &   \ul{0.244}   &   \ul{0.244}   &   \ul{0.244}   &   \ul{0.359}   &   \ul{0.359}   &   \ul{0.359} \\
protoroles-existed\_during  &   \ul{-0.011}   &   \ul{-0.011}   &   \ul{-0.011}   &   \textbf{0.194}   &   \textbf{0.194}   &   \textbf{0.194}   &   \ul{0.074}   &   \ul{0.074}   &   \ul{0.074}   &   \ul{0.146}   &   \ul{0.146}   &   \ul{0.146} \\
protoroles-exists\_as\_physical  &   \ul{0.086}   &   \ul{0.086}   &   \ul{0.086}   &   \textbf{0.304}   &   \textbf{0.304}   &   \textbf{0.304}   &   \ul{-0.045}   &   \ul{-0.045}   &   \ul{-0.045}   &   \ul{0.079}   &   \ul{0.079}   &   \ul{0.079} \\ 
protoroles-instigation  &   \ul{0.343}   &   \ul{0.343}   &   \ul{0.343}   &   \textbf{0.425}   &   \textbf{0.425}   &   \textbf{0.425}   &   \ul{0.301}   &   \ul{0.301}   &   \ul{0.301}   &   \ul{0.406}   &   \ul{0.406}   &   \ul{0.406} \\
protoroles-location\_of\_event  &   \ul{-0.059}   &   \ul{-0.059}   &   \ul{-0.059}   &   \ul{-0.485}   &   \ul{-0.485}   &   \ul{-0.485}   &   \ul{-0.292}   &   \ul{-0.292}   &   \ul{-0.292}   &   \textbf{0.184}   &   \textbf{0.184}   &   \textbf{0.184} \\
protoroles-makes\_physical\_contact  &   \ul{-0.035}   &   \ul{-0.035}   &   \ul{-0.035}   &   \textbf{0.323}   &   \textbf{0.323}   &   \textbf{0.323}   &   \ul{0.149}   &   \ul{0.149}   &   \ul{0.149}   &   \ul{0.146}   &   \ul{0.146}   &   \ul{0.146} \\
protoroles-partitive  &   \ul{0.079}   &   \ul{0.079}   &   \ul{0.079}   &   \ul{0.059}   &   \ul{0.059}   &   \ul{0.059}   &   \ul{0.048}   &   \ul{0.048}   &   \ul{0.048}   &   \textbf{0.121}   &   \textbf{0.121}   &   \textbf{0.121} \\
protoroles-predicate\_changed\_argument  &   \textbf{0.342}   &   \textbf{0.342}   &   \textbf{0.342}   &   \ul{0.099}   &   \ul{0.099}   &   \ul{0.099}   &   \ul{-0.078}   &   \ul{-0.078}   &   \ul{-0.078}   &   \ul{-0.175}   &   \ul{-0.175}   &   \ul{-0.175} \\
protoroles-sentient  &   \ul{0.678}   &   \ul{0.678}   &   \ul{0.678}   &   \textbf{0.748}   &   \textbf{0.748}   &   \textbf{0.748}   &   \ul{0.614}   &   \ul{0.614}   &   \ul{0.614}   &   \ul{0.724}   &   \ul{0.724}   &   \ul{0.724} \\
protoroles-stationary  &   \ul{-0.1}   &   \ul{-0.1}   &   \ul{-0.1}   &   \textbf{0.04}   &   \textbf{0.04}   &   \textbf{0.04}   &   \ul{-0.154}   &   \ul{-0.154}   &   \ul{-0.154}   &   \ul{-0.0}   &   \ul{-0.0}   &   \ul{-0.0} \\
protoroles-volition  &   \ul{0.606}   &   \ul{0.606}   &   \ul{0.606}   &   \textbf{0.685}   &   \textbf{0.685}   &   \textbf{0.685}   &   \ul{0.56}   &   \ul{0.56}   &   \ul{0.56}   &   \ul{0.677}   &   \ul{0.677}   &   \ul{0.677} \\
protoroles-was\_for\_benefit  &   \ul{0.327}   &   \ul{0.327}   &   \ul{0.327}   &   \ul{0.36}   &   \ul{0.36}   &   \ul{0.36}   &   \ul{0.259}   &   \ul{0.259}   &   \ul{0.259}   &   \textbf{0.369}   &   \textbf{0.369}   &   \textbf{0.369} \\
protoroles-was\_used  &   \ul{0.058}   &   \ul{0.058}   &   \ul{0.058}   &   \ul{0.031}   &   \ul{0.031}   &   \ul{0.031}   &   \ul{0.079}   &   \ul{0.079}   &   \ul{0.079}   &   \textbf{0.105}   &   \textbf{0.105}   &   \textbf{0.105} \\
relation-classification  &   \textbf{0.362}   &   \textbf{0.362}   &   \textbf{0.362}   &   \ul{0.281}   &   \ul{0.281}   &   \ul{0.281}   &   \ul{0.244}   &   \ul{0.244}   &   \ul{0.244}   &  0.27  &  0.27  &   \ul{0.271} \\
senteval-coordination\_inversion  &  0.623  &   \ul{0.649}   &  0.555  &   \ul{0.628}   &  0.614  &  0.554  &   \ul{0.715}   &  0.639  &  0.583  &   \textbf{0.721}   &  0.698  &  0.614 \\
senteval-odd\_man\_out  &   \ul{0.622}   &  0.612  &  0.575  &   \ul{0.63}   &  0.584  &  0.552  &   \textbf{0.736}   &  0.7  &  0.619  &   \ul{0.73}   &  0.714  &  0.661 \\
senteval-past\_present  &   \textbf{0.885}   &  0.883  &  0.844  &   \ul{0.873}   &  0.86  &  0.818  &   \ul{0.87}   &  0.725  &  0.716  &   \ul{0.862}   &  0.843  &  0.716 \\
senteval-word\_content  &   \textbf{0.084}   &  0.04  &  0.02  &   \ul{0.02}   &  0.0  &  0.014  &   \ul{0.006}   &  0.0  &  0.002  &   \ul{0.006}   &  0.004  &  0.005 \\
sentiment-sentence  &   \textbf{0.425}   &  0.237  &  0.133  &   \ul{0.234}   &  0.082  &  0.113  &  0.163  &   \ul{0.184}   &  0.07  &   \ul{0.269}   &  0.061  &  0.084 \\
srl  &   \ul{0.114}   &   \ul{0.114}   &   \ul{0.114}   &   \ul{0.105}   &   \ul{0.105}   &   \ul{0.105}   &   \ul{0.118}   &   \ul{0.118}   &   \ul{0.118}   &  0.125  &   \textbf{0.126}   &   \textbf{0.126} \\
synonym-antonym-hard  &   \ul{0.608}   &  0.588  &  0.5  &   \ul{0.559}   &  0.514  &  0.458  &   \textbf{0.63}   &  0.591  &  0.494  &   \ul{0.606}   &  0.59  &  0.499 \\
time  &   \ul{0.105}   &   \ul{0.105}   &   \ul{0.105}   &   \textbf{0.106}   &   \textbf{0.106}   &   \textbf{0.106}   &   \ul{0.099}   &   \ul{0.099}   &   \ul{0.099}   &   \ul{0.098}   &   \ul{0.098}   &   \ul{0.098} \\
wordsense  &   \textbf{0.104}   &   \textbf{0.104}   &   \textbf{0.104}   &   \ul{0.084}   &   \ul{0.084}   &   \ul{0.084}   &   \ul{0.071}   &   \ul{0.071}   &   \ul{0.071}   &   \ul{0.086}   &   \ul{0.086}   &   \ul{0.086} \\
zorro-quantifiers-existential\_there  &   \ul{0.773}   &  0.678  &  0.636  &   \textbf{0.905}   &  0.438  &  0.621  &  0.638  &   \ul{0.74}   &  0.581  &   \ul{0.803}   &  0.789  &  0.716 \\
zorro-quantifiers-superlative  &   \ul{0.981}   &  0.76  &  0.616  &   \textbf{0.997}   &  0.425  &  0.634  &   \ul{0.973}   &  0.652  &  0.673  &   \ul{0.995}   &  0.93  &  0.722 \\ \hline 
 average & 0.463  &  0.449  &  0.398  &  0.468  &  0.386  &  0.405  &  0.436  &  0.373  &  0.374  &   \textbf{0.474}   &  0.46  &  0.409  \\ \hline
    \caption{FlashHolmes semantic tasks results. In \textbf{bold} are the overall best results, and \ul{underlined} are the results for the best performing variation for each transformer.}
    \label{tab:semantics}
\end{longtable}

\newpage

\begin{longtable}{l|ccc|ccc|ccc|ccc} \\
    Data & \multicolumn{3}{c}{BERT} & \multicolumn{3}{c}{RoBERTa} & \multicolumn{3}{c}{DeBERTa}  & \multicolumn{3}{c}{Electra} \\ 
  &  S-avg  &  CLS  &  Rand  &  S-avg  &  CLS  &  Rand &  S-avg  &  CLS  &  Rand &  S-avg  &  CLS  &  Rand \\ \hline \endhead
bridging-edge  &   \ul{0.563}   &   \ul{0.563}   &   \ul{0.563}   &   \ul{0.519}   &   \ul{0.519}   &   \ul{0.519}   &   \ul{0.574}   &   \ul{0.574}   &   \ul{0.574}   &   \textbf{0.614}   &   \textbf{0.614}   &   \textbf{0.614} \\
bridging-sentence  &  0.4  &  0.4  &   \textbf{0.424}   &   \ul{0.4}   &   \ul{0.4}   &   \ul{0.4}   &  0.4  &  0.4  &   \textbf{0.424}   &  0.4  &  0.4  &   \ul{0.405} \\
coref  &   \ul{0.695}   &   \ul{0.695}   &   \ul{0.695}   &   \ul{0.725}   &   \ul{0.725}   &   \ul{0.725}   &   \ul{0.718}   &   \ul{0.718}   &   \ul{0.718}   &   \textbf{0.735}   &   \textbf{0.735}   &   \textbf{0.735} \\
discourse-connective  &  0.073  &  0.068  &   \ul{0.082}   &   \ul{0.055}   &  0.023  &  0.047  &   \ul{0.064}   &  0.023  &  0.056  &  0.064  &   \textbf{0.093}   &  0.071 \\
gum-rst-edu-count  &   \ul{0.031}   &   \ul{0.031}   &   \ul{0.031}   &   \textbf{0.034}   &   \textbf{0.034}   &   \textbf{0.034}   &   \ul{0.014}   &   \ul{0.014}   &   \ul{0.014}   &   \ul{0.02}   &   \ul{0.02}   &   \ul{0.02} \\
gum-rst-edu-depth  &   \ul{0.105}   &   \ul{0.105}   &   \ul{0.105}   &   \ul{0.146}   &   \ul{0.146}   &   \ul{0.146}   &   \ul{0.091}   &   \ul{0.091}   &   \ul{0.091}   &   \textbf{0.151}   &   \textbf{0.151}   &   \textbf{0.151} \\
gum-rst-edu-distance  &   \ul{0.067}   &   \ul{0.067}   &   \ul{0.067}   &   \textbf{0.13}   &   \textbf{0.13}   &   \textbf{0.13}   &   \ul{0.038}   &   \ul{0.038}   &   \ul{0.038}   &   \ul{0.052}   &   \ul{0.052}   &   \ul{0.052} \\
gum-rst-edu-relation  &   \ul{0.067}   &   \ul{0.067}   &   \ul{0.067}   &   \ul{0.059}   &   \ul{0.059}   &   \ul{0.059}   &   \ul{0.072}   &   \ul{0.072}   &   \ul{0.072}   &   \textbf{0.076}   &   \textbf{0.076}   &   \textbf{0.076} \\
gum-rst-edu-relation-group  &   \ul{0.071}   &   \ul{0.071}   &   \ul{0.071}   &   \ul{0.061}   &   \ul{0.061}   &   \ul{0.061}   &   \textbf{0.109}   &   \textbf{0.109}   &   \textbf{0.109}   &   \ul{0.092}   &   \ul{0.092}   &   \ul{0.092} \\
gum-rst-edu-successively  &   \textbf{0.486}   &   \textbf{0.486}   &   \textbf{0.486}   &   \ul{0.483}   &   \ul{0.483}   &   \ul{0.483}   &   \ul{0.483}   &   \ul{0.483}   &   \ul{0.483}   &   \ul{0.479}   &   \ul{0.479}   &   \ul{0.479} \\
gum-rst-edu-type  &   \ul{0.664}   &   \ul{0.664}   &   \ul{0.664}   &   \ul{0.658}   &   \ul{0.658}   &   \ul{0.658}   &   \textbf{0.689}   &   \textbf{0.689}   &   \textbf{0.689}   &   \ul{0.68}   &   \ul{0.68}   &   \ul{0.68} \\
next-sentence-prediction  &  0.429  &   \textbf{0.615}   &  0.429  &   \ul{0.429}   &   \ul{0.429}   &   \ul{0.429}   &   \ul{0.502}   &  0.429  &  0.43  &  0.429  &   \ul{0.497}   &  0.429 \\
ordering  &   \ul{0.865}   &   \ul{0.865}   &   \ul{0.865}   &   \textbf{0.923}   &   \textbf{0.923}   &   \textbf{0.923}   &   \ul{0.727}   &   \ul{0.727}   &   \ul{0.727}   &   \ul{0.798}   &   \ul{0.798}   &   \ul{0.798} \\ \hline
    averages & 0.347  &   \textbf{0.361}   &  0.350  &  0.356  &  0.353  &  0.355  &  0.345  &  0.336  &  0.340  &  0.353  &   \textbf{0.361}   &  0.354 \\ \hline
    \caption{FlashHolmes discourse tasks results. In \textbf{bold} are the overall best results, and \ul{underlined} are the results for the best performing variation for each transformer.}
    \label{tab:discourse}
\end{longtable}

\newpage

\begin{longtable}{l|ccc|ccc|ccc|ccc} \\
    Data & \multicolumn{3}{c}{BERT} & \multicolumn{3}{c}{RoBERTa} & \multicolumn{3}{c}{DeBERTa}  & \multicolumn{3}{c}{Electra} \\ 
  &  S-avg  &  CLS  &  Rand  &  S-avg  &  CLS  &  Rand &  S-avg  &  CLS  &  Rand &  S-avg  &  CLS  &  Rand \\ \hline \endhead
SemAntoNeg  &   \ul{0.4}   &   \ul{0.4}   &   \ul{0.4}   &   \ul{0.4}   &   \ul{0.4}   &   \ul{0.4}   &  0.4  &  0.4  &   \textbf{0.408}   &   \ul{0.4}   &   \ul{0.4}   &   \ul{0.4} \\
bioscope-negation-span-classify  &   \textbf{0.979}   &   \textbf{0.979}   &   \textbf{0.979}   &   \ul{0.969}   &   \ul{0.969}   &   \ul{0.969}   &   \ul{0.971}   &   \ul{0.971}   &   \ul{0.971}   &   \ul{0.97}   &   \ul{0.97}   &   \ul{0.97} \\
bioscope-negation-span-classify  &   \textbf{0.601}   &   \textbf{0.601}   &   \textbf{0.601}   &   \ul{0.589}   &   \ul{0.589}   &   \ul{0.589}   &   \ul{0.58}   &   \ul{0.58}   &   \ul{0.58}   &   \ul{0.577}   &   \ul{0.577}   &   \ul{0.577} \\
bioscope-org-negation  &  0.212  &  0.207  &   \ul{0.266}   &  0.299  &  0.294  &   \ul{0.338}   &  0.279  &  0.259  &   \textbf{0.417}   &  0.228  &  0.207  &   \ul{0.276} \\
fuse-negation-span-classify  &   \ul{0.919}   &   \ul{0.919}   &   \ul{0.919}   &   \ul{0.495}   &   \ul{0.495}   &   \ul{0.495}   &   \ul{0.917}   &   \ul{0.917}   &   \ul{0.917}   &   \textbf{0.942}   &   \textbf{0.942}   &   \textbf{0.942} \\
fuse-negation-span-correspondence  &   \ul{0.48}   &   \ul{0.48}   &   \ul{0.48}   &   \ul{0.33}   &   \ul{0.33}   &   \ul{0.33}   &   \textbf{0.512}   &   \textbf{0.512}   &   \textbf{0.512}   &   \ul{0.458}   &   \ul{0.458}   &   \ul{0.458} \\
fuse-org-negation  &  0.2  &  0.196  &   \ul{0.239}   &  0.28  &  0.29  &   \textbf{0.329}   &  0.203  &  0.196  &   \ul{0.275}   &  0.211  &  0.196  &   \ul{0.267} \\
olmpics-antonym\_synonym\_negation  &   \ul{0.515}   &  0.507  &  0.46  &  0.455  &  0.396  &   \ul{0.483}   &   \textbf{0.749}   &  0.392  &  0.477  &  0.665  &   \ul{0.703}   &  0.48 \\
olmpics-coffee\_cats\_quantifiers  &   \ul{0.444}   &   \ul{0.444}   &   \ul{0.444}   &   \ul{0.444}   &   \ul{0.444}   &   \ul{0.444}   &  0.444  &  0.444  &   \ul{0.453}   &  0.444  &  0.444  &   \textbf{0.46} \\
olmpics-composition\_v2  &  0.4  &  0.4  &   \textbf{0.409}   &  0.4  &  0.4  &   \ul{0.407}   &  0.4  &  0.4  &   \ul{0.404}   &  0.4  &  0.4  &   \ul{0.407} \\
olmpics-compositional\_comparison  &  0.4  &  0.4  &   \ul{0.404}   &  0.4  &  0.4  &   \ul{0.407}   &  0.4  &  0.4  &   \textbf{0.47}   &  0.4  &  0.4  &   \ul{0.434} \\
olmpics-conjunction\_filt4  &  0.4  &  0.4  &   \ul{0.403}   &  0.4  &  0.4  &   \textbf{0.412}   &   \ul{0.4}   &   \ul{0.4}   &   \ul{0.4}   &  0.4  &  0.4  &   \ul{0.403} \\
olmpics-hypernym\_conjunction  &  0.401  &   \ul{0.408}   &  0.4  &   \ul{0.4}   &   \ul{0.4}   &   \ul{0.4}   &   \textbf{0.42}   &  0.4  &  0.402  &  0.4  &   \ul{0.41}   &  0.404 \\
olmpics-number\_comparison\_age\_compare\_masked  &  0.404  &  0.476  &   \ul{0.487}   &  0.38  &  0.372  &   \ul{0.457}   &   \textbf{0.919}   &  0.333  &  0.433  &   \ul{0.638}   &  0.603  &  0.459 \\
olmpics-size\_comparison  &   \ul{0.486}   &  0.423  &  0.479  &   \ul{0.427}   &  0.347  &  0.421  &   \textbf{0.556}   &  0.394  &  0.51  &  0.429  &  0.434  &   \ul{0.447} \\
sherlock-negation  &  0.434  &  0.434  &   \ul{0.436}   &   \ul{0.434}   &   \ul{0.434}   &   \ul{0.434}   &   \ul{0.435}   &  0.434  &   \ul{0.435}   &  0.434  &  0.434  &   \textbf{0.437} \\
speculation-org  &  0.196  &  0.195  &   \ul{0.224}   &  0.265  &   \textbf{0.319}   &  0.306  &  0.195  &  0.195  &   \ul{0.23}   &  0.198  &  0.195  &   \ul{0.256} \\
speculation-span-classify  &   \ul{0.514}   &   \ul{0.514}   &   \ul{0.514}   &   \ul{0.47}   &   \ul{0.47}   &   \ul{0.47}   &   \textbf{0.526}   &   \textbf{0.526}   &   \textbf{0.526}   &   \ul{0.505}   &   \ul{0.505}   &   \ul{0.505} \\
speculation-span-correspondence  &   \ul{0.465}   &   \ul{0.465}   &   \ul{0.465}   &   \ul{0.417}   &   \ul{0.417}   &   \ul{0.417}   &   \ul{0.453}   &   \ul{0.453}   &   \ul{0.453}   &   \textbf{0.475}   &   \textbf{0.475}   &   \textbf{0.475}  \\ \hline      
  average & 0.466  &  0.466  &  0.474  &  0.434  &  0.430  &  0.448  &   \textbf{0.514}   &  0.453  &  0.488  &  0.483  &  0.482  &  0.477  \\ \hline
    \caption{FlashHolmes reasoning tasks results. In \textbf{bold} are the overall best results, and \ul{underlined} are the results for the best performing variation for each transformer.}
    \label{tab:reasoning}
\end{longtable}

\end{landscape}

\section{Probing for structure}

\begin{table*}[h]
    \centering
    \begin{tabular}{rrrrrr}
  train on & test on &   BERT      &    RoBERTa    &    DeBERTa    & Electra \\ \hline
      CLS  &   CLS & 0.896 (0.088) & 0.789 (0.027) & 0.227 (0.058) & 0.955 (0.006) \\
           &   AVG & 0.910 (0.078) & 0.793 (0.026) & 0.130 (0.025) & 0.971 (0.003) \\
           &  RAND & 0.919 (0.070) & 0.792 (0.023) & 0.139 (0.019) & 0.966 (0.002) \\ \cline{2-6}
      AVG  &   CLS & 1.000 (0.000) & 0.943 (0.013) & 0.174 (0.020) & 0.999 (0.001) \\
           &   AVG & 0.999 (0.001) & 0.936 (0.017) & 0.325 (0.087) & 0.997 (0.001) \\
           &  RAND & 1.000 (0.000) & 0.939 (0.018) & 0.327 (0.096) & 0.999 (0.001) \\ \cline{2-6}
      RAND &   CLS & 0.998 (0.001) & 0.888 (0.009) & 0.163 (0.023) & 0.999 (0.001) \\
           &   AVG & 0.998 (0.002) & 0.895 (0.004) & 0.233 (0.048) & 0.998 (0.001) \\
           &  RAND & 0.997 (0.003) & 0.886 (0.005) & 0.221 (0.048) & 0.997 (0.003) \\ \hline
    \end{tabular}
    \caption{Detailed results on detecting chunk structure in sentence embeddings. Averaged F1 scores (standard deviation) over three runs.}
    \label{tab:sent_str}
\end{table*}

\end{document}